\documentclass[11pt]{article}
\pdfoutput = 1

\usepackage[dvips]{graphicx}
\usepackage{amssymb,amsmath,amsthm,color}
\usepackage{url}
\usepackage{natbib}
\bibpunct{(}{)}{;}{a}{,}{,}
 \usepackage[colorlinks,
            linkcolor=blue,
            citecolor=blue,
            urlcolor=magenta,
            linktocpage,
            plainpages=false]{hyperref}

\renewcommand\epsilon\varepsilon 

\newcommand\al{\alpha} 
\newcommand\be{\beta} 
\newcommand\eps{\epsilon} 

\newcommand\EE{\mathbb E} 
\newcommand\NN{\mathbb{N}} 
\newcommand\RR{\mathbb{R}} 
\newcommand\CC{\mathbb{C}} 

\newcommand\tn{\theta_{n}} 
\newcommand\tnp{\theta_{n+1}} 
\newcommand\tnm{\theta_{n-1}} 
\newcommand\pn{\phi_{n}} 

\newcommand{\BEAS}{\begin{eqnarray*}}
\newcommand{\EEAS}{\end{eqnarray*}}
\newcommand{\BEA}{\begin{eqnarray}}
\newcommand{\EEA}{\end{eqnarray}}
\newcommand{\BEQ}{\begin{equation}}
\newcommand{\EEQ}{\end{equation}}
\newcommand{\BIT}{\begin{itemize}}
\newcommand{\EIT}{\end{itemize}}
\newcommand{\BNUM}{\begin{enumerate}}
\newcommand{\ENUM}{\end{enumerate}}
\newcommand{\BA}{\begin{array}}
\newcommand{\EA}{\end{array}}

\newcommand{\Tr}{\mathop{ \rm tr}}
\newcommand{\tr}{\mathop{ \rm tr}}

\newcommand{\idm}{I}

\newcommand{\mysec}[1]{Section~\ref{sec:#1}}
\newcommand{\eq}[1]{Eq.~(\ref{eq:#1})}
\newcommand{\myfig}[1]{Figure~\ref{fig:#1}}

\newtheorem{lemma}{Lemma}
\newtheorem{theorem}{Theorem}
\newtheorem{proposition}{Proposition}
\newtheorem{corollary}{Corollary}

\title{From Averaging to Acceleration, There is Only a Step-size}

\author{
Nicolas Flammarion and Francis Bach\\
INRIA - Sierra project-team\\
D\'epartement d'Informatique de l'Ecole Normale Sup\'erieure \\
Paris, France \\
\texttt{nicolas.flammarion@ens.fr}, \texttt{francis.bach@ens.fr}
}
\begin{document}

\maketitle

\begin{abstract}
We show that accelerated gradient descent, averaged gradient descent and the heavy-ball method for non-strongly-convex problems may be reformulated as   constant parameter second-order difference equation algorithms, where stability of the system is equivalent to convergence at rate $O(1/n^2)$, where $n$ is the number of iterations. We provide a detailed analysis of the eigenvalues of the corresponding linear dynamical system, showing various oscillatory and non-oscillatory behaviors, together with a sharp stability result with explicit constants. We also consider the situation where noisy gradients are available, where we extend our general convergence result, which suggests an alternative algorithm (i.e., with different step sizes) that exhibits the good aspects of both averaging and acceleration.
 \end{abstract}

\section{Introduction}

Many problems in machine learning are naturally cast as convex optimization problems over a Euclidean space; for supervised learning this includes least-squares regression, logistic regression, and the support vector machine. Faced with large amounts of data, practitioners often favor first-order techniques based on gradient descent, leading to algorithms with many cheap iterations. For smooth problems, two extensions of gradient descent have had important theoretical and practical impacts: acceleration and averaging.

Acceleration techniques date back to~\citet{n1} and have their roots in momentum techniques and conjugate gradient~\citep{MR1099605}. For convex problems, with an appropriately weighted momentum term which requires to store two iterates, \citet{n1} showed that the traditional convergence rate of $O(1/n)$ for the function values after $n$ iterations of gradient descent goes down to $O(1/n^2)$ for accelerated gradient descent, such a rate being optimal among first-order techniques that can access only sequences of gradients~\citep{MR2142598}. Like conjugate gradient methods for solving linear systems, these methods are however more sensitive to noise in the gradients; that is, to preserve their improved convergence rates, significantly less noise may be tolerated~\citep{as,schmidt:inria-00618152,MR3232608}.

Averaging techniques which consist in replacing the iterates by the average of all iterates have also been thoroughly considered, either because they sometimes lead to simpler proofs, or because they lead to improved behavior. In the noiseless case where gradients are exactly available, they do not improve the convergence rate in the convex case; worse, for strongly-convex problems, they are not linearly convergent while regular gradient descent is. Their main advantage comes with random unbiased gradients, where it has been shown that they lead to better convergence rates than the unaveraged counterparts, in particular because they allow larger step-sizes~\citep{pj,bm2}. For example, for least-squares regression with stochastic gradients, they lead to convergence rates of $O(1/n)$, even in the non-strongly convex case~\citep{bm1}.

In this paper, we show that for quadratic problems, both averaging and acceleration are two instances of the same second-order finite difference equation, with different step-sizes. They may thus be analyzed jointly, together with a non-strongly convex version of the heavy-ball method~\citep[][Section 3.2]{MR1099605}. In presence of random zero-mean noise on the gradients, this joint analysis allows to design a novel intermediate algorithm that exhibits the good aspects of both acceleration (quick forgetting of initial conditions) and averaging (robustness to noise).

In this paper, we make the following contributions:

\vspace*{-.15cm}

\begin{list}{\labelitemi}{\leftmargin=1.7em}
   \addtolength{\itemsep}{-.215\baselineskip}

\item[--] We show in \mysec{soa} that accelerated gradient descent, averaged gradient descent and the heavy-ball method for non-strongly-convex problems may be reformulated as   constant parameter second-order difference equation algorithms, where stability of the system is equivalent to convergence at rate $O(1/n^2)$.
 \item[--] In \mysec{conv}, we provide a detailed analysis of the eigenvalues of the corresponding linear dynamical system, showing various oscillatory and non-oscillatory behaviors, together with a sharp stability result with explicit constants.
 \item[--] In \mysec{noise}, we consider the situation where noisy gradients are available, where we extend our general convergence result, which suggests an alternative algorithm (i.e., with different step sizes) that exhibits the good aspects of both averaging and acceleration.
 \item[--] In \mysec{simu}, we illustrate our results with simulations on synthetic examples.
\end{list}

\section{Second-Order Iterative Algorithms for Quadratic Functions}\label{sec:soa}

Throughout this paper, we consider minimizing a convex quadratic function $f:\RR^d \to \RR$ defined as:
\begin{equation}
 f (\theta) =  \frac{1}{2}\langle \theta, H \theta\rangle-\langle q,\theta\rangle,
\end{equation}
with $H\in\RR^{d\times d}$ a symmetric positive semi-definite   matrix and $q\in \RR^d$. Without loss of generality, $H$ is assumed invertible (by projecting onto the orthogonal of its null space), though its eigenvalues could be arbitrarily small.
The solution is known to be $\theta_*=H^{-1}q$, but  the inverse of the Hessian is often too expensive to compute when $d$ is large.
The excess cost function may be simply expressed as \mbox{$f(\theta_n)-f(\theta_*)=\frac{1}{2}\langle \theta_n-\theta_*,H(\theta_n-\theta_*)\rangle$}.

\subsection{Second-order algorithms}

\vspace*{-.1256cm}

In this paper we study second-order iterative algorithms of the form:
\begin{equation}\label{eq:theta}
 \tnp=A_n\tn+B_n\tnm+c_n,
\end{equation}
started with $\theta_1 =  \theta_0$ in $\RR^d$, with $A_n\in\RR^{d\times d}$, $B_n\in\RR^{d\times d}$ and $c_n\in\RR^{d}$ for all $n\in\NN^*$. We impose the natural restriction that  the optimum $\theta_*$ is a stationary point of this recursion, that is, for all $n\in\NN^*$:
\begin{equation}\label{optimum-centering} \tag{$\theta_*$-stationarity}
 \theta_*=A_n\theta_*+B_n\theta_*+c_n.
\end{equation}
 By letting $\phi_n=\theta_n-\theta_*$ we then have $
  \phi_{n+1}=A_n\phi_n+B_n\phi_{n-1}$,
started from $\phi_0= \phi_1  = \theta_0-\theta_* $. Thus, we restrict our problem to the study of the convergence of an iterative system to $0$.

In connection with accelerated methods, we are interested in algorithms for which $f(\theta_n)-f(\theta_*) = \frac{1}{2}\langle \phi_n,H\phi_n\rangle$  converges to $0$ at a speed of $O\left(1/n^2\right)$.
Within this context we impose that $A_n$ and $B_n$ have the form~:
\begin{equation}
\label{n-scalability}\tag{n-scalability}
 A_n=\frac{n}{n+1}A\text{  and } B_n=\frac{n-1}{n+1}B  \text{\quad  $\forall n\in\NN$} \text{ with $A,B\in\RR^{d\times d}$} .
\end{equation}
By letting $\eta_n= n \phi_n = n ( \theta_n-\theta_*)$, we can now study the simple iterative system with \emph{constant} terms $ \eta_{n+1}=A\eta_n+B\eta_{n-1}$, started at $\eta_0=0$ and $\eta_1=\theta_0-\theta_*$.
Showing that the function values remain bounded, we directly have the convergence of $f(\theta_n)$ to $f(\theta_*)$ at the speed $O\left(1/n^2\right)$.
Thus the \ref{n-scalability} property allows to switch from a convergence problem to a stability problem.

For feasibility concerns the method can only access $H$ through matrix-vector products.
Therefore $A$ and $B$ should be polynomials in $H$ and $c$ a polynomial in $H$ times $q$, if possible of low degree. The following theorem clarifies the general form of iterative systems which share these three properties (see proof in Appendix~\ref{app:soa}).
\begin{theorem}
\label{theo:1}
 Let $(P_n,Q_n,R_n)\in (\RR[X])^3$ for all $n\in\NN$, be a sequence of polynomials. If the iterative algorithm defined by \eq{theta} with $A_n=P_n(H)$, $B_n=Q_n(H)$ and $c_n=R(H)q$ satisfies the  \ref{optimum-centering} and  \ref{n-scalability} properties,
 there are polynomials  $(\bar A,\bar B)\in(\RR[X])^2$ such that:
 \begin{eqnarray*}
    A_n&=&2\frac{n}{n+1}\bigg(I-\bigg(\frac{\bar A(H)+\Bar B(H)}{2}\bigg)H\bigg),\\[-.1cm]
    B_n&=&-\frac{n-1}{n+1}\bigg(I-\bar B(H) H\bigg)\ \  \mbox{ and }  \ \
    c_n = \bigg(\frac{n\bar A(H)+\bar B(H)}{n+1} \bigg) q.
 \end{eqnarray*}
\end{theorem}
Note that our result prevents $A_n$ and $B_n$ from being zero, thus requiring the algorithm to strictly be of second order.  This illustrates the fact that first-order algorithms as gradient descent do not have the convergence rate in $O(1/n^2)$.

We now restrict our class of algorithms to lowest possible order polynomials, that is, $\bar A=\alpha I$ and $\bar B=\beta I$ with $(\alpha,\beta)\in\RR^{2}$, which correspond to the fewest matrix-vector products per iteration, leading to the \emph{constant-coefficient} recursion for $\eta_n=n\pn = n (\theta_n - \theta_\ast)$:
\begin{equation}\label{eq:eta}
 \eta_{n+1}=\left(I-\alpha H\right)\eta_n+\left(I-\beta H\right)\left(\eta_n-\eta_{n-1}\right).
\end{equation}

\paragraph{Expression with gradients of $f$.}
The recursion in \eq{eta} may be written with gradients of $f$ in multiple ways. In order to preserve the parallel with accelerated techniques, we rewrite it as:
\BEQ
\label{eq:thetafinal}
\theta_{n+1} =  \frac{2n}{n+1} \theta_n - \frac{n-1}{n+1} \theta_{n-1}
- \frac{n \alpha + \beta}{n+1} f' \bigg(
\frac{n(\alpha+\beta)}{n\alpha + \beta} \theta_n - \frac{(n-1)\beta}{n\alpha + \beta} \theta_{n-1}
\bigg).
\EEQ
It may be interpreted as a modified gradient recursion with two potentially different affine (i.e., with coefficients that sum to one) combinations of the two past iterates.
This reformulation will also be crucial when using noisy gradients.
The allowed values for $(\alpha,\beta)\in\RR^2$ will be determined in the following sections.

\subsection{Examples}

\vspace*{-.1256cm}

\paragraph{Averaged gradient descent.}

We consider averaged  gradient descent (referred to from now on as ``Av-GD'') \citep{pj} with step-size $\gamma\in\RR$ defined by:
$$
  \psi_{n+1} =  \psi_n -\gamma f'(\psi_n), \ \ \  \
   \theta_{n+1} = \frac{1}{n+1}\sum_{i=1}^{n+1}\psi_{i}.
$$
When computing the average online as $
     \tnp=\tn+\frac{1}{n+1}(\psi_{n+1}-\tn)$ and seeing the average as the main iterate, the algorithm becomes (see proof  in Appendix~\ref{sec:agdacc}):
 $$ \tnp = \frac{2n}{n+1} \theta_n - \frac{n-1}{n+1} \theta_{n-1}-\frac{\gamma}{n+1} f' \big(
n\theta_n - (n-1)\theta_{n-1}
\big).
 $$
This corresponds to \eq{thetafinal} with $\alpha = 0$ and $\beta = \gamma$.

\paragraph{Accelerated gradient descent.}
We consider  the accelerated gradient  descent (referred to from now on as ``Acc-GD'')  \citep{n1} with step-sizes $(\gamma,\delta_n)\in\RR^2$ :
$$
\tnp = \omega_n-\gamma f'(\omega_n),  \ \ \ \ \
\omega_n = \tn+\delta_n(\tn-\tnm).
$$
For smooth optimization the accelerated literature \citep{MR2142598, bt} uses the step-size $\delta_n=1-\frac{3}{n+1}$ and their results are not valid  for bigger step-size $\delta_n$.
However $\delta_n=1-\frac{2}{n+1}$ is compatible with the framework of \cite{lan} and is more convenient for our set-up.
This corresponds to \eq{thetafinal} with $\alpha = \gamma $ and $\beta = \gamma$. Note that accelerated techniques are more generally applicable, e.g., to composite optimization with smooth functions~\citep{MR3071865,bt}.

\paragraph{Heavy ball.}
We consider  the    heavy-ball algorithm (referred to from now on as ``HB'') \citep{Polyak1964} with step-sizes $(\gamma,\delta_n)\in\RR^2$ :
\begin{equation*}
 \tnp=\tn-\gamma f'(\tn)+\delta_n(\tn-\tnm),
\end{equation*}
when $\delta_n=1-\frac{2}{n+1}$.
We note that typically $\delta_n$ is constant for strongly-convex problems.
This corresponds to \eq{thetafinal} with $\alpha = \gamma $ and $\beta = 0$.

\section{Convergence with Noiseless Gradients}\label{sec:conv}

\vspace*{-.1255cm}

We study the convergence of the iterates defined by:
$
  \eta_{n+1}=\left(I-\alpha H\right)\eta_n+\left(I-\beta H\right)\left(\eta_n-\eta_{n-1}\right)$.
This is a second-order iterative system with constant coefficients that it is standard to cast in a linear framework \citep[see, e.g.,][]{ort}. We may rewrite it as:
$$
 \Theta_{n}=F\Theta_{n-1}, \  \mbox{ with }
 \Theta_n=\begin{pmatrix}
                 \eta_n\\
                 \eta_{n-1}
                \end{pmatrix} \mbox{ and }  F=\begin{pmatrix}
                 2I-\left(\alpha+\beta \right)H &\beta H-I \\
                I &0
                \end{pmatrix}\in\RR^{2d\times2d}.
$$
Thus $\Theta_n = F^n \Theta_0$. Following~\citet{o2013adaptive}, if we consider an eigenvalue decomposition of $H$, i.e., $H=P \text{Diag}(h) P^{\top}$ with $P$ an orthogonal matrix and $(h_i)$ the eigenvalues of $H$, sorted in decreasing order:
 $h_{d}=L\geq h_{d-1}\geq \cdots\geq h_{2}\geq h_{1}=\mu > 0$, then   \eq{eta} may be rewritten as:
\begin{equation}
 P^{\top}\eta_{n+1}=\left(I-\al \text{Diag}\left(h\right)\right)P^{\top}\eta_n+\left(I-\be \text{Diag}\left(h\right)\right)\big(P^{\top}\eta_n-P^{\top}\eta_{n-1}\big).
\end{equation}
Thus there is no interaction between the different eigenspaces and we may consider, for the analysis only, $d$ different recursions with $\eta_n^i= p_i^{\top}\eta_{n}$,   $i\in\{1,...,d\}$, where $p_{i}\in \RR^{d}$ is the $i$-th column of~$P$:
\begin{equation}
 \eta^i_{n+1}=\left(1-\al h_i\right)\eta^i_n+\left(1-\beta h_i\right)\left(\eta^i_n-\eta^i_{n-1}\right).
\end{equation}

\subsection{Characteristic polynomial and eigenvalues}

\vspace*{-.1256cm}

In this section, we consider a fixed $i \in \{1,\dots,d\}$ and study the stability in the corresponding eigenspace.
This linear dynamical system may be analyzed by studying the eigenvalues of the $2\times 2$-matrix $F_{i}=\begin{pmatrix}
                 2-(\alpha+\beta) h_{i}&\beta  h_{i}-1 \\
                1 &0
                \end{pmatrix}$.
These eigenvalues are the roots of its characteristic polynomial which is:
$$ \! \det( XI-F_{i}) = \det\left( X\left(X-2+(\alpha+\beta) h_{i}\right)+1-\beta h_{i} \right)
= X^2-2X\Big(1-\Big(\frac{\alpha+\beta}{2}\Big) h_{i}\Big)+1-\beta h_{i}.
$$
To compute the roots of the second-order polynomial, we compute its  reduced discriminant:
\begin{equation*}
 \Delta_i=\Big(1-\Big(\frac{\alpha + \beta}{2}\Big) h_i\Big)^2-1+\beta h_i =  h_i\Big(\Big(\frac{\alpha+\beta}{2}\Big)^2 h_i-\alpha \Big).
\end{equation*}
Depending on the sign of the discriminant $\Delta_i$,
there will be two real distinct eigenvalues \mbox{($\Delta_i>0)$},
two complex conjugate eigenvalues \mbox{($\Delta_i<0)$} or a single real eigenvalue \mbox{($\Delta_i=0)$}.

We will now study the sign of $\Delta_i$.
In each different case, we will determine under what conditions on $\alpha$  and $\beta$ the modulus of the eigenvalues is less than one, which means that the iterates $(\eta^i_{n})_n$
remain bounded and the iterates $(\theta_{n})_n$ converge to $\theta_{*}$. We may then compute function values as  $f(\theta_n)-f(\theta_\ast)=\frac{1}{2n^2}\sum_{i=1}^d (\eta_n^i)^2h_i =  \frac{1}{2}\sum_{i=1}^d (\phi_n^i)^2h_i$.

\begin{figure}
\centering

\vspace*{-.25cm}

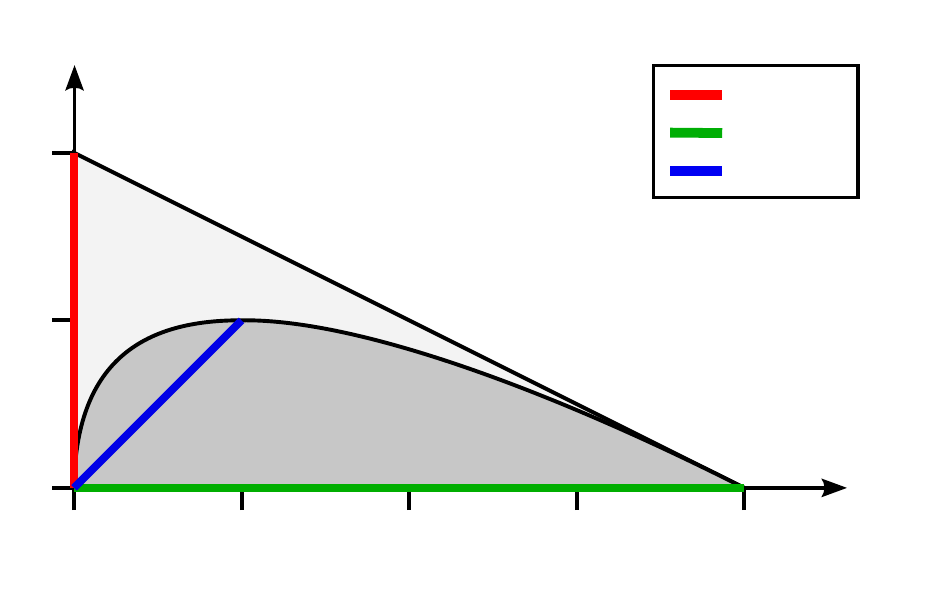

\vspace*{-.75cm}

\caption{Area of stability of the algorithm, with the three traditional algorithms represented. In the interior of the triangle, the convergence is linear.}
\label{fig:stab}
\end{figure}

The various regimes are summarized in \myfig{stab}: there is a triangle of values of $(\alpha h_i,\beta h_i)$ for which the algorithm remains stable (i.e., the iterates $(\eta_n)_n$ do not diverge), with either complex or real eigenvalues. In the following lemmas (see proof in Appendix~\ref{app:appB}), we provide a detailed analysis that leads to  \myfig{stab}.

\begin{lemma}[Real eigenvalues]\label{lemma:stabreal}
The discriminant $\Delta_i$ is strictly positive and the algorithm is stable if and only if
$$
 \alpha \geq 0 , \ \ \ \
 \alpha+2\beta \leq  4/h_i , \ \ \ \
 \alpha+\beta >  2\sqrt{\alpha/h_i}
 .$$
We then have two real roots
$
 r_i^\pm=r_i\pm\sqrt{\Delta_i}$,  with $r_i=1-(\frac{\alpha+\beta}{2})h_i$. Moreover,
 we have:
\begin{equation}
 ({\phi_n^i})^2h_i=\frac{{(\phi_1^i)}^2h_i}{4n^2}\frac{\left[(r_i+\sqrt{\Delta_i})^n-(r_i-\sqrt{\Delta_i})^n\right]^2}{\Delta_i}.
\end{equation}
\end{lemma}
Therefore, for real eigenvalues, $( ({\phi_n^i}) ^2 h_i)_n$ will converge to $0$ at a speed of $O(1/n^{2})$ however the constant $\Delta_i$ may be arbitrarily small (and thus the scaling factor arbitrarily large). Furthermore we have linear convergence if the inequalities in the lemmas are strict.

\begin{lemma}[Complex eigenvalues]\label{lemma:stabcom}
The discriminant $\Delta_i$ is  stricly negative and  the algorithm is stable if and only if
$$
 \alpha \geq 0, \ \ \ \
 \beta \geq  0 , \ \ \ \
\alpha+\beta < \sqrt{\alpha/h_i}.
$$
We then have two complex conjugate eigenvalues:
$  r_i^\pm=r_i\pm \sqrt{-1} \sqrt{-\Delta_i}$.
Moreover, we have:
\begin{equation}
 {(\phi_n^i)}^2h_i=\frac{({\phi_1^i})^2}{n^2}\frac{\sin^2(\omega_i n)}{\big(\al-(\frac{\al+\be}{2})^2 h_i\big)}\rho^{2n}.
 \end{equation}
with $\rho_i=\sqrt{ 1-\beta h_i  }$, and $\omega_i$ defined through $\sin (\omega_i)=\sqrt{-\Delta_i}/\rho_i$ and $\cos(\omega_i)=r_i/\rho_i$.
\end{lemma}
Therefore, for complex eigenvalues, there is a linear convergence if the inequalities in the lemma are strict. Moreover, $( ({\phi_n^i}) ^2 h_i)_n$  oscillates to $0$ at a speed of $O(1/n^{2})$ even if $h_{i}$ is arbitrarily small.

 \paragraph{Coalescing eigenvalues.} When the discriminant goes to zero in the explicit formulas of the real and complex cases, both the denominator and numerator of $( ({\phi_n^i})^2h_i)_n$ will go to zero.
In the limit case, when the discriminant is equal to zero, we  will have a double real eigenvalue. This happens for $\beta=2\sqrt{\alpha/h_i}-\alpha$.
 Then the eigenvalue is
$
  r_i=1-\sqrt{\alpha h_i}
$,
and the algorithm is stable for $0<\al <  {4}/{h_i}$, we then have $
 ({\phi_n^i})^2h_i=h_i({\phi_1^i})^2(1-\sqrt{\alpha h_i})^{2(n-1)}$.
This can be obtained by letting $\Delta_{i}$ goes to $0$ in the real and complex cases (see also Appendix~\ref{app:coal}).

\paragraph{Summary.}

To conclude the iterate $(\eta_n^i)_n = ( n ( \theta_n^i - \theta_\ast^i) )_n$ will be stable for $\alpha\in[0,4/h_i]$ and $\beta \in [0, 2/h_i-\alpha/2]$.
According to the values of $\alpha$ and $\beta$ this iterate will have a different behavior.
In the complex case, the roots are complex conjugate with magnitude $\sqrt{1-\beta h_{i}}$.
Thus, when $\beta>0$, $(\eta_{n}^i)_n$ will converge to $0$, oscillating, at rate $\sqrt{1-\beta h_{i}}$.
In the real case, the two roots are real and distinct. However the product of the two roots is equal to  $\sqrt{1-\beta h_{i}}$,
thus one will have a higher magnitude and  $(\eta_{n}^i)_n$ will converges to $0$ at rate higher than in the complex case (as long as $\alpha$ and $\beta$ belong to the interior of the stability region).

Finally, for a given quadratic function $f$, all the $d$ iterates $(\eta_n^i)_n$ should be bounded, therefore we must have \mbox{$\alpha\in[0,4/L]$} and \mbox{$\beta \in [0, 2/L-\alpha/2]$}.  Then, depending on the value of $h_i$, some eigenvalues may be complex or real.

\subsection{Classical examples}

\vspace*{-.1256cm}

For particular choices of $\al$ and $\be$, displayed in \myfig{stab}, the eigenvalues are either all real or all complex, as shown in the table below.

\begin{center}
\begin{tabular}{|l|l|l|l|}
  \hline
   & Av-GD & Acc-GD & Heavy ball \\
  \hline\hline
  $\alpha$ &$ 0$ & $\gamma $ & $\gamma$\\

  $\beta$ & $\gamma$ & $\gamma$&$0 $  \\

  $\Delta_i$&$(\gamma h_i)^2$&$-\gamma h_i(1-\gamma h_i)$&$-\gamma h_i(1-\frac{\gamma h_i}{4})$\\

  $r_i^\pm$& $1$, $1-\gamma h_i$&$\sqrt{1-\gamma h_i}e^{\pm i \omega_i}$&$e^{\pm i\omega_i}$ \\

  $\cos(\omega_i)$& & $\sqrt{1-\gamma h_i}$ &$ {1-\frac{\gamma}{2} h_i}$\\

 $\rho_{i}$& & $\sqrt{1-\gamma h_i}$ &$ 1$\\
  \hline \end{tabular}
\end{center}
Averaged gradient descent loses linear convergence for strongly-convex problems, because $r_i^+=1$ for all eigensubspaces. Similarly, the heavy ball method is not adaptive to strong convexity because $\rho_i = 1$. However, accelerated gradient descent, although designed for non-strongly-convex problems, is adaptive because $\rho_i = \sqrt{1-\gamma h_i}$ depends on $h_i$ while $\alpha$ and $\beta$ do not. These last two algorithms have an oscillatory behavior which can be observed in practice and has been already studied~\citep{su2014differential}.

 Note that all the classical methods choose step-sizes $\alpha$ and $\beta$ either having all the eigenvalues real either complex; whereas we will see in \mysec{noise}, that it is  significant to combine both behaviors in presence of noise.

\subsection{General bound}\label{sec:generalbound}

\vspace*{-.1256cm}

Even if the exact formulas in Lemmas~\ref{lemma:stabreal} and \ref{lemma:stabcom} are computable, they are not easily interpretable.
In particular when the two roots become close, the denominator will go to zero, which prevents from bounding them easily.
When we further restrict the domain of $(\alpha,\beta)$, we can always bound the iterate by the general bound (see proof in Appendix~\ref{app:bounddet}):
\begin{theorem}\label{theo:bounddet}
 For $\alpha\leq 1/h_i$ and $0 \leq \beta\leq 2 /h_i-\alpha$, we have
\begin{equation}
  (\eta_n^i)^2\leq\min\bigg\{\frac{2(\eta_1^i)^2}{{\alpha h_i}},\frac{8(\eta_1^i)^2 n}{{(\alpha+{\beta})h_i}},\frac{16(\eta_1^i)^2 }{(\alpha+\beta)^2h_i^2} \bigg\}.
\end{equation}
\end{theorem}
These bounds are shown by dividing the set of $(\alpha,\beta)$ in three regions where we obtain specific bounds. They do not depend on the regime of the eigenvalues (complex or real); this   enables us to get the following general bound on the function values, our main result for the deterministic case.
\begin{corollary}
\label{cor:det}
For $\alpha\leq 1/L$ and $0 \leq \beta\leq 2 /L-\alpha$:
 \begin{equation}\label{eq:laz}
   f(\theta_n)-f(\theta_*)  \leq \min \bigg\{\frac{{\Vert \theta_0-\theta_*\Vert}^2}{\alpha n^2}, \frac{4{\Vert\theta_0-\theta_*\Vert}^2}{(\alpha+\beta)n}\bigg\}.
 \end{equation}
\end{corollary}
We can make the following observations:

\vspace*{-.15cm}

\begin{list}{\labelitemi}{\leftmargin=1.7em}
   \addtolength{\itemsep}{-.215\baselineskip}

\item[--]  The first bound $\frac{{\Vert \theta_0-\theta_*\Vert}^2}{\alpha n^2}$ corresponds to the traditional acceleration result, and is only relevant for $\alpha > 0$ (that is, for Nesterov acceleration and the heavy-ball method, but not for averaging). We recover the traditional convergence rate of second-order methods for quadratic functions in the singular case, such as conjugate gradient~\citep[][Section~6.1]{MR1099605}.

\item[--] While the result above focuses on function values, like most results in the non-strongly convex case, the distance to optimum $\| \theta_n - \theta_\ast\|^2$ typically does not go to zero (although it remains bounded in our situation).

\item[--] When $\alpha = 0$ (averaged gradient descent), then the second bound $ \frac{4{\Vert\theta_0-\theta_*\Vert}^2}{(\alpha+\beta)n}$ provides a convergence rate  of $O(1/n)$ if no assumption is made regarding the starting point $\theta_0$, while the last bound of Theorem \ref{theo:bounddet} would lead to a bound $\frac{{8\Vert H^{-1/2}(\theta_0-\theta_*)\Vert}^2}{(\alpha+\beta)^2n^2}$, that is a rate of $O(1/n^2)$, only for some starting points.
\item[--] As shown in Appendix~\ref{app:lb} by exhibiting explicit sequences of quadratic functions, the inverse dependence in $\alpha n^2$ and $(\alpha + \beta) n$ in \eq{laz} is not improvable.
\end{list}

\section{Quadratic Optimization with Additive Noise }\label{sec:noise}

\vspace*{-.1255cm}

In many practical situations, the gradient of $f$ is not available for the recursion in \eq{thetafinal}, but only a noisy version. In this paper, we only consider additive uncorrelated noise with finite variance.

\begin{figure}
\centering

\vspace*{-.25cm}

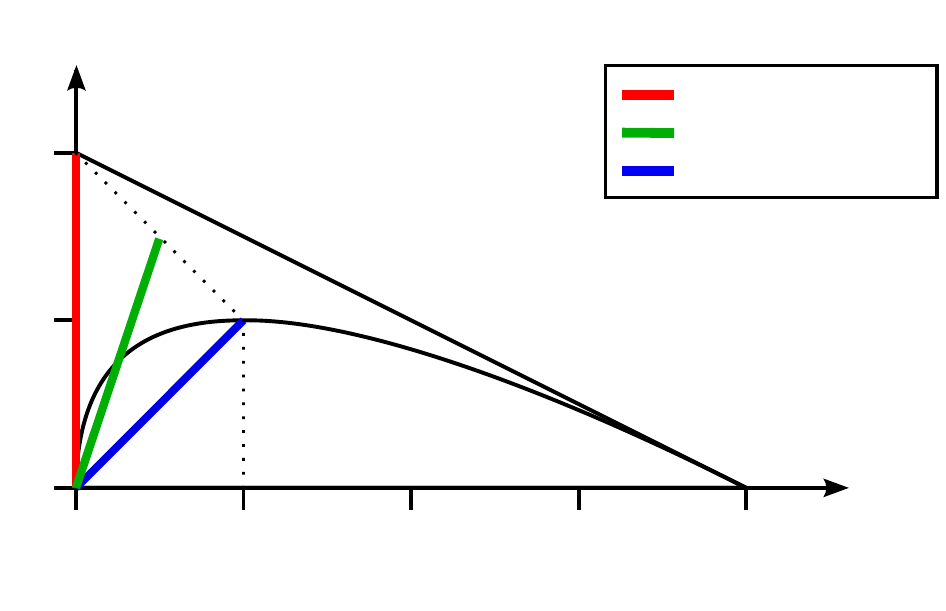

\vspace*{-.75cm}

\caption{Trade-off between averaged and accelerated methods for noisy gradients.}
\end{figure}

\subsection{Stochastic difference equation}

\vspace*{-.1256cm}

We now assume that the true gradient is not available and  we rather have  access to a noisy oracle for the gradient of $f$. In \eq{thetafinal}, we assume that the oracle outputs  a noisy gradient
$f' \big(
\frac{n(\alpha+\beta)}{n\alpha + \beta} \theta_n - \frac{(n-1)\beta}{n\alpha + \beta} \theta_{n-1}
\big)  - \varepsilon_{n+1}$. The noise $(\eps_n)$ is assumed to be uncorrelated zero-mean with bounded covariance, i.e., \mbox{$\EE[\eps_{n}\otimes\eps_{m}]= 0$} for all $n\neq m$ and  $\EE[\eps_{n}\otimes\eps_{n}]\preccurlyeq C$, where $A\preccurlyeq B$ means that $B-A$ is positive semi-definite.

 For quadratic functions, for the reduced variable   $\eta_n=n\phi_n =  n (\theta_n - \theta_\ast)$, we get:
\begin{equation}
 \eta_{n+1}=(I-\al H)\eta_n +(I-\be H)(\eta_n-\eta_{n-1})+[n\al+\be] \eps_{n+1}.
\end{equation}
Note that algorithms with $\alpha\neq0$ will have an important level of noise because of the term $n\alpha\eps_{n+1}$.
We denote by $\xi_{n+1}=\begin{pmatrix}
              [n\al+\be] \eps_{n+1}\\
              0
             \end{pmatrix}$
and we now have the recursion:
\begin{equation}
 \Theta_{n+1}=F\Theta_n+\xi_{n+1},
\end{equation}
which is a standard noisy linear dynamical system \citep[see, e.g.,][]{randynsys} with uncorrelated noise process $(\xi_n)$. We may thus express $\Theta_n$ directly as
$
\Theta_n = F^n \Theta_0 + \sum_{k=1}^n F^{n-k} \xi_k,
$
and its expected second-order moment as,
$
\EE \big( \Theta_n \Theta_n \big)^\top =  F^n \Theta_0 \Theta_0^\top (F^n)^\top +
 \sum_{k=1}^n F^{n-k} \EE \big( \xi_k \xi_k^\top \big) (F^{n-k} )^\top.
$
In order to obtain the expected excess cost function, we simply need to compute
$
\tr \begin{pmatrix}
            0 & H \\
              0& 0
                           \end{pmatrix} \EE \big( \Theta_n \Theta_n \big)^\top
$, which thus decomposes as a term that only depends on initial conditions (which is exactly the one computed and studied in \mysec{generalbound}), and a new term that depends on the noise.

\subsection{Convergence result}

\vspace*{-.1256cm}

For a quadratic function $f$ with arbitrarily  small eigenvalues and   uncorrelated noise with finite covariance, we obtain the following convergence result (see proof in Appendix~\ref{app:noise}); since we will allow the parameters $\alpha$ and $\beta$ to depend on the time we stop the algorithm, we introduce the horizon~$N$:
\begin{theorem}[Convergence rates with noisy gradients]\label{cor:gennoise}
With $\EE[\eps_n\otimes\eps_n]=C$ for all $n\in \NN$,  for $\alpha\leq\frac{1}{L}$ and $0\leq \beta\leq \frac{2}{L}-\alpha$. Then for any $N \in \NN$, we have:
\begin{multline}
 \EE f(\theta_N)-f(\theta_*) \leq
 \\ \min \bigg\{\frac{{\Vert\theta_0-\theta_*\Vert}^2}{\alpha N^2}+ \frac{(\alpha N+\beta)^{2}}{\alpha N} \Tr (C), \frac{4{\Vert\theta_0-\theta_*\Vert}^2}{(\alpha+\beta)N}+\frac{4(\alpha N+\beta)^{2}}{\alpha+\beta} \Tr (C)\bigg\}.
 \end{multline}
\end{theorem}
We can make the following observations:

\vspace*{-.15cm}

\begin{list}{\labelitemi}{\leftmargin=1.7em}
   \addtolength{\itemsep}{-.215\baselineskip}

\item[--] Although we only provide an upper-bound, the proof technique relies on direct moment computations in each eigensubspace with few inequalities, and we conjecture that the scalings with respect to $n$ are tight.

\item[--] For $\alpha=0$ and  $\beta = 1/L$ (which corresponds to averaged gradient descent), the second bound leads to  $\frac{4{ L \Vert\theta_0-\theta_*\Vert}^2}{ N}+ \frac{4 \Tr (C)}{L}$, which is bounded but not converging to zero. We recover a result from~\citet[Theorem 1]{bm2}.
\item[--] For $\alpha = \beta = 1/L$ (which corresponds to Nesterov's acceleration), the first bound leads to
$\frac{ { L \Vert\theta_0-\theta_*\Vert}^2}{ N^2}+ \frac{(N+1) \Tr (C)}{L}$, and our bound suggests that the algorithm diverges, which we have observed in our experiments in Appendix \ref{app:det}.

\item[--] For $\alpha = 0$ and $\beta = 1/L \sqrt{N}$, the second bound leads to
$\frac{ {4 L \Vert\theta_0-\theta_*\Vert}^2}{ \sqrt{N}}+ \frac{4 \Tr (C)}{L \sqrt{N}}$, and
we recover the traditional rate of $1/\sqrt{N}$ for stochastic gradient in the non-strongly-convex case.

\item[--]  When the values of the bias and the variance are known we can choose $\alpha$ and $\beta$ such that the trade-off between the bias and the variance is optimal in our bound, as the following corrollary shows. Note that in the bound below, taking a non zero $\beta$ enables the bias term to be adaptive to hidden strong-convexity.
\end{list}
\begin{corollary}\label{cor:oa}
 For $\alpha=\min\left\{\frac{\Vert\theta_0-\theta_*\Vert}{2\sqrt{\tr C} N^{3/2}},1/L\right\}$ and $\beta \in [0, \min\{ N \alpha, 1/L \}]$, we have:
 \begin{equation*}
 \EE f(\theta_N)-f(\theta_*) \leq \frac{2L \Vert\theta_0-\theta_*\Vert^2 }{N^2} + \frac{4\sqrt{\tr C}\Vert\theta_0-\theta_*\Vert }{\sqrt{N}}
  .
 \end{equation*}
\end{corollary}

\subsection{Structured noise and least-square regression}
\label{sec:ls}

\vspace*{-.1256cm}

When only the noise total variance $\tr (C)$ is considered, as shown in \mysec{related}, Corollary \ref{cor:oa} recover existing (more general) results. Our framework however leads to improved result for \emph{structured noise processes} frequent in machine learning, in particular in  least-squares regression which we now consider but this goes beyond \citep[see, e.g.][]{bm1}.

Assume we observe independent and identically distributed pairs $(x_n,y_n)\in \RR^d \times \RR$ and we want to minimize the expected loss
\mbox{$f(\theta)=\frac{1}{2}\EE [ (y_n - \langle \theta,x_n\rangle)^2]$}.
We denote by $H=\EE( x_n\otimes x_n)$ the covariance matrix which is assumed invertible. The global minimum of $f$ is attained at $\theta_*\in\RR^d$ defined as before and we denote by $r_n=y_n-\langle\theta_*,x_n\rangle  $ the statistical noise, which we assume bounded by $\sigma$. We have $\EE [r_n x_n]=0$. In an online setting, we observe the gradient
$
(x_n\otimes x_n) (\theta-\theta_*)- r_n x_n
$, whose expectation is the gradient $f'(\theta)$. This corresponds to a noise in the gradient of
$\varepsilon_n = (H-x_n\otimes x_n  ) (\theta-\theta_*)+ r_n x_n$. Given $\theta$, if the data $(x_n,y_n)$ are almost surely bounded, the covariance matrix of this noise is bounded by a constant times $H$. This suggests to characterize the noise convergence by $ \tr ( C H^{-1})$, which is bounded even though $H$ has arbitrarily small eigenvalues.

However, our result will not apply to stochastic gradient  descent (SGD) for least-squares, because of the term $(H - x_n\otimes x_n ) (\theta-\theta_*)$ which depends on $\theta$, but to a ``semi-stochastic'' recursion where the noisy gradient is $H(\theta-\theta_*)- r_n x_n$, with a noise process $\varepsilon_n = r_n x_n$, which is such that $\EE [\eps_n\otimes\eps_n]\preccurlyeq\sigma^2H $, and has been used by \cite{bm2} and \cite{db} to prove results on regular stochastic gradient descent. We conjecture that our algorithm (and results) also applies in the regular SGD case, and we provide encouraging experiments in \mysec{simu}.

For this particular structured noise we can take advantage of a large $\beta$:
\begin{theorem}[Convergence rates with structured noisy gradients]\label{theo:gen}
Let $\alpha\leq\frac{1}{L}$ and $0\leq \beta\leq \frac{3}{2L}-\frac{\alpha}{2}$. For any $N \in \NN$,
$ \EE f(\theta_N)-f(\theta_*) $ is upper-bounded by:
\begin{equation}\label{eq:theogen}
\min \left\{\frac{\Vert \theta_0-\theta_*\Vert^2}{N^2\alpha}+\frac{(\alpha N+\beta)^2}{\alpha\beta N^2}\tr ( C H^{-1}), \frac{4L\Vert\theta_0-\theta_*\Vert^2}{(\alpha+\beta)N}+\frac{8(\alpha N+\beta)^2\tr ( C H^{-1})}{(\alpha+\beta)^2N} \right\}.
\end{equation}
\end{theorem}
We can make the following observations:

\vspace*{-.15cm}

\begin{list}{\labelitemi}{\leftmargin=1.7em}
   \addtolength{\itemsep}{-.215\baselineskip}

\item[--] For $\alpha=0$ and  $\beta = 1/L$ (which corresponds to averaged gradient descent), the second bound leads to  $\frac{4{ L \Vert\theta_0-\theta_*\Vert}^2}{ N}+ \frac{8 \tr ( C H^{-1})}{N}$. We recover a result from~\citet[Theorem 1]{bm1}. Note that when $C \preccurlyeq \sigma^2 H$,  $\tr ( C H^{-1}) \leqslant \sigma^2 d$.
\item[--] For $\alpha = \beta = 1/L$ (which corresponds to Nesterov's acceleration), the first bound leads to
$\frac{ { L \Vert\theta_0-\theta_*\Vert}^2}{ N^2}+  \tr ( C H^{-1})$, which is bounded but not converging to zero (as opposed to the the unstructured noise where the algorithm may diverge).
\item[--] For $\alpha=1/(L N^a)$ with $0\leq a \leq 1$ and $ \beta= 1/L$,  the first bound leads to $\frac{L \Vert \theta_0-\theta_*\Vert^2}{N^{2-a}}+\frac{ \tr ( C H^{-1})}{N^a}$.
We thus obtain an explicit bias-variance trade-off by changing the value of $a$.

 \item[--]  When the values of the bias and the variance are known we can choose $\alpha$ and $\beta$ with an optimized trade-off, as the following corrollary shows:

\end{list}
\begin{corollary}\label{cor:ob}
 For $\alpha=\min\left\{\frac{\Vert\theta_0-\theta_*\Vert}{\sqrt{L\tr(C H^{-1})} N},1/L\right\}$ and $\beta=\min\left\{N\alpha,1/L\right\}$ we have:
 \begin{equation}
 \EE f(\theta_N)-f(\theta_*) \leq \max\bigg\{\frac{5\tr(C H^{-1})}{N},\frac{5\sqrt{ \tr(C H^{-1}) L}\Vert\theta_0-\theta_*\Vert }{N},\frac{2\Vert\theta_0-\theta_*\Vert^2L}{N^2}\bigg\}.
 \end{equation}
\end{corollary}

\subsection{Related work}
\label{sec:related}

\vspace*{-.1256cm}

 \paragraph{Acceleration and noisy gradients.} Several authors \citep{lan,sage,xiao} have shown that using a step-size proportional to $1/N^{3/2}$ accelerated methods with noisy gradients lead to the same
 convergence rate of $O\big(\frac{L\Vert \theta_0-\theta_*\Vert^2}{N^2}+\frac{\Vert \theta_0-\theta_*\Vert \sqrt{\tr(C)}}{\sqrt{N}}\big)$ than in Corollary~\ref{cor:oa}, for smooth functions. Thus, for unstructured noise, our analysis provides insights in the behavior of second-order algorithms, without improving  bounds. We get significant improvements for structured noises.

\paragraph{Least-squares regression.} When the noise is structured as in least-square regression and more generally in linear supervised learning, \cite{bm2} have shown that using averaged stochastic gradient descent with constant step-size leads to the convergence rate of
$O\big(\frac{L\Vert \theta_0-\theta_0 \Vert^2}{N}+\frac{\sigma^2 d}{N}\big)$. It has been highlighted by~\citet{defossez2014constant} that the bias term  $\frac{L\Vert\theta_0-\theta_*\Vert^2}{N}$ may often be the dominant one in practice. Our result in Corollary~\ref{cor:ob} leads to an improved bias term in $O(1/N^2)$ with the price of a potentially slightly worse constant in the variance term. However, with optimal constants in Corollary \ref{cor:ob}, the new algorithm is always an improvement over averaged stochastic gradient descent in all situations. If constants are unknown, we may use $\alpha=1/(L N^a)$ with $0\leq a \leq 1$ and $ \beta= 1/L$ and we choose $a$ depending on the emphasis we want to put on bias or variance.

\paragraph{Minimax convergence rates.}
For noisy quadratic problems, the convergence rate nicely decomposes into two terms, a bias term which corresponds to the noiseless problem and the variance term which corresponds to a problem started at $\theta_\ast$. For each of these two terms, lower bounds are known. For the bias term, if $N \leq d$, then the lower bound is, up to constants, $ L\| \theta_0 - \theta_\ast\|^2 / N^2$~\citep[][Theorem~2.1.7]{MR2142598}. For the variance term, for the general noisy gradient situation, we show in Appendix~\ref{app:minimax} that for $N \leq d$,   it is  $  {(\tr C)}/({L \sqrt{N}})$, while for least-squares regression, it is $\sigma^2 d / N$~\citep{tsybakov2003optimal}. Thus, for the two situations, we attain the two lower bounds \emph{simultaneously} for situations where respectively $L \| \theta_0 - \theta_\ast\|^2 \leq (\tr C)/L$ and $L \| \theta_0 - \theta_\ast\|^2 \leq d \sigma^2$. It remains an open problem to achieve the two minimax terms in all situations.

 \paragraph{Other algorithms  as special cases.}

We also note as shown in Appendix~\ref{app:comparison} that in the special case of quadratic functions, the algorithms of  \citet{lan,sage,xiao} could be unified into our framework (although they have significantly different formulations and justifications in the smooth case).

\section{Experiments}
\label{sec:simu}

\vspace*{-.1255cm}

In this section, we illustrate our theoretical results on synthetic examples.
We consider a matrix~$H$  that has random eigenvectors and eigenvalues $1/k^m$, for $k=1,\dots,d$ and $m\in\NN$.
We take a random optimum $\theta_*$ and a random starting point $\theta_0$ such that $r=\Vert \theta_0-\theta_{*}\Vert = 1$ (unless otherwise specified). In Appendix~\ref{app:det}, we illustrate the noiseless results of \mysec{conv}, in particular the oscillatory behaviors and the influence of all eigenvalues, as well as unstructured noisy gradients. In this section, we focus on noisy gradients with structured noise (as described in \mysec{ls}), where our new algorithms show significant improvements.

We compare our algorithm to other stochastic accelerated algorithms, that is,   AC-SA \citep{lan}, SAGE \citep{sage} and Acc-RDA \citep{xiao} which are presented in Appendix~\ref{app:comparison}.
For all these algorithms (and ours) we take the optimal step-sizes defined in these papers.
We show results averaged over 10 replications.

\begin{figure}

  \vspace*{-.25cm}

    \hspace*{.5cm}
\begin{minipage}[c]{.4 \linewidth}
\includegraphics[width=\linewidth]{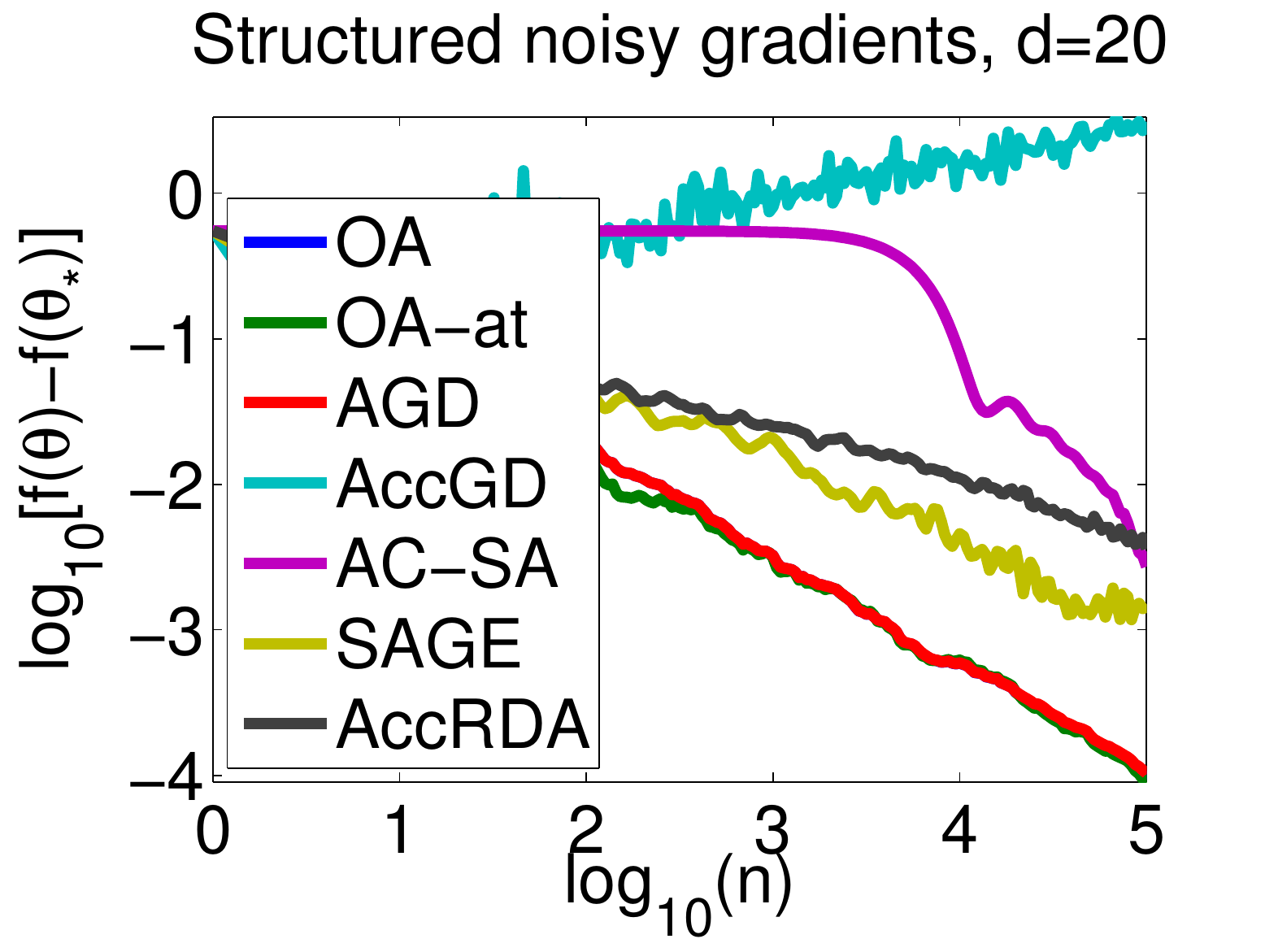}
   \end{minipage}
   \hspace*{.08\linewidth}
   \begin{minipage}[c]{.4 \linewidth}
\includegraphics[width=\linewidth]{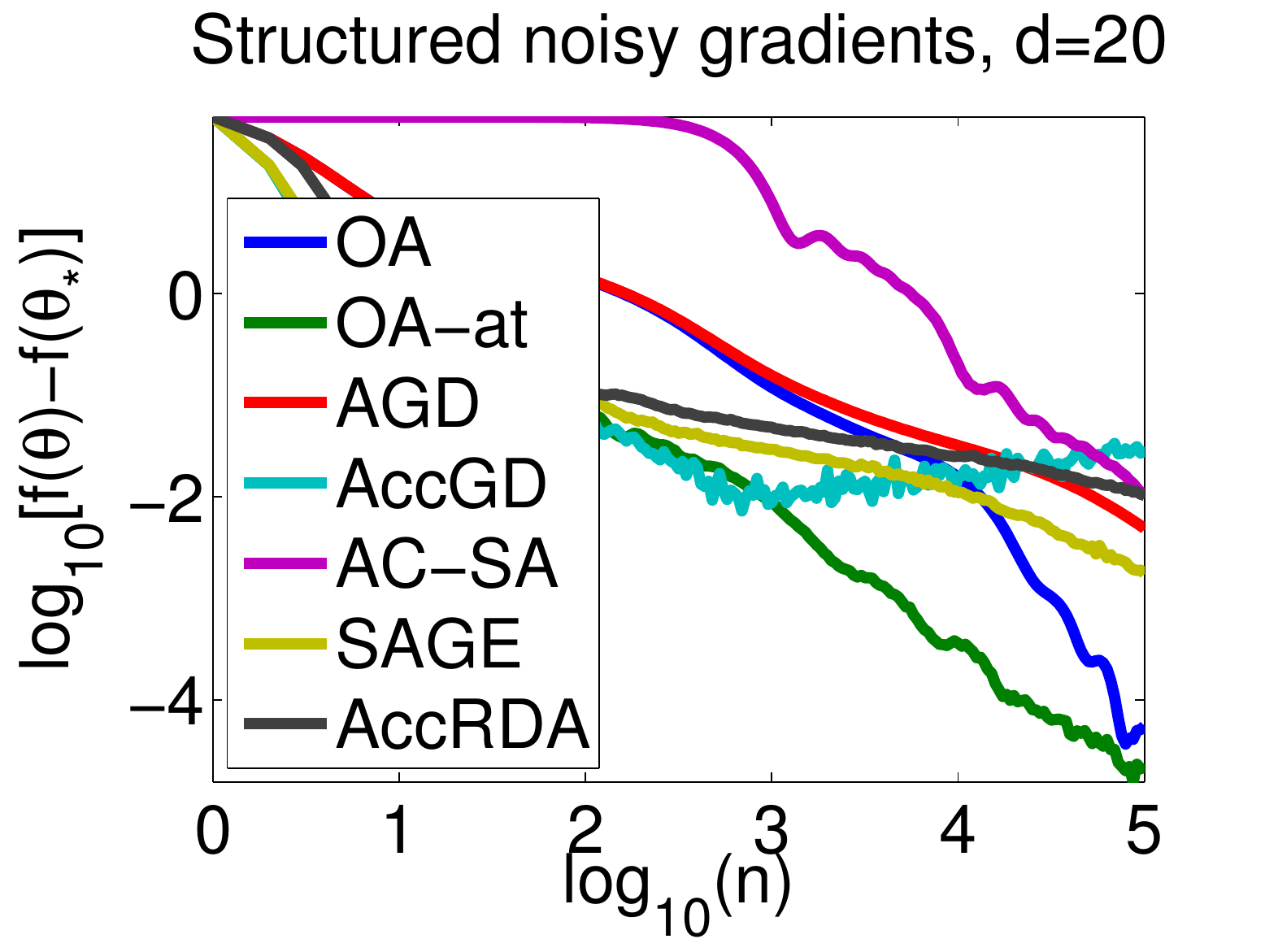}
      \end{minipage}

      \vspace*{-.3cm}

    \caption{Quadratic optimization with regression noise. Left $\sigma=1$, $r=1$. Right $\sigma=0.1$, $r=10$.}
   \label{fig:plot6}
\end{figure}

\paragraph{Homoscedastic noise.} We first consider an i.i.d.~zero mean noise whose covariance  matrix is proportional to $H$. We also consider a variant of our algorithm with an any-time step-size function of $n$ rather than $N$ (for which we currently have no proof of convergence).
In Figure \ref{fig:plot6}, we take into account two different set-ups.
In the left plot,  the variance dominates the bias (with $r = \| \theta_0 - \theta_\ast\|=\sigma$).
We  see that (a) Acc-GD does not converge to the optimum but does not   diverge either, (b) Av-GD and our algorithms achieve the optimal rate of convergence of $O(\sigma^{2}d/n)$, whereas (c)
 other accelerated algorithms only converge at rate $O(1/\sqrt{n})$.
In the right plot, the bias dominates the variance ($r=10$ and $\sigma=0.1$).
In this situation our algorithm outperforms all  others.

\begin{figure}

\hspace*{.5cm}
\begin{minipage}[c]{.4  \linewidth}
\includegraphics[width=\linewidth]{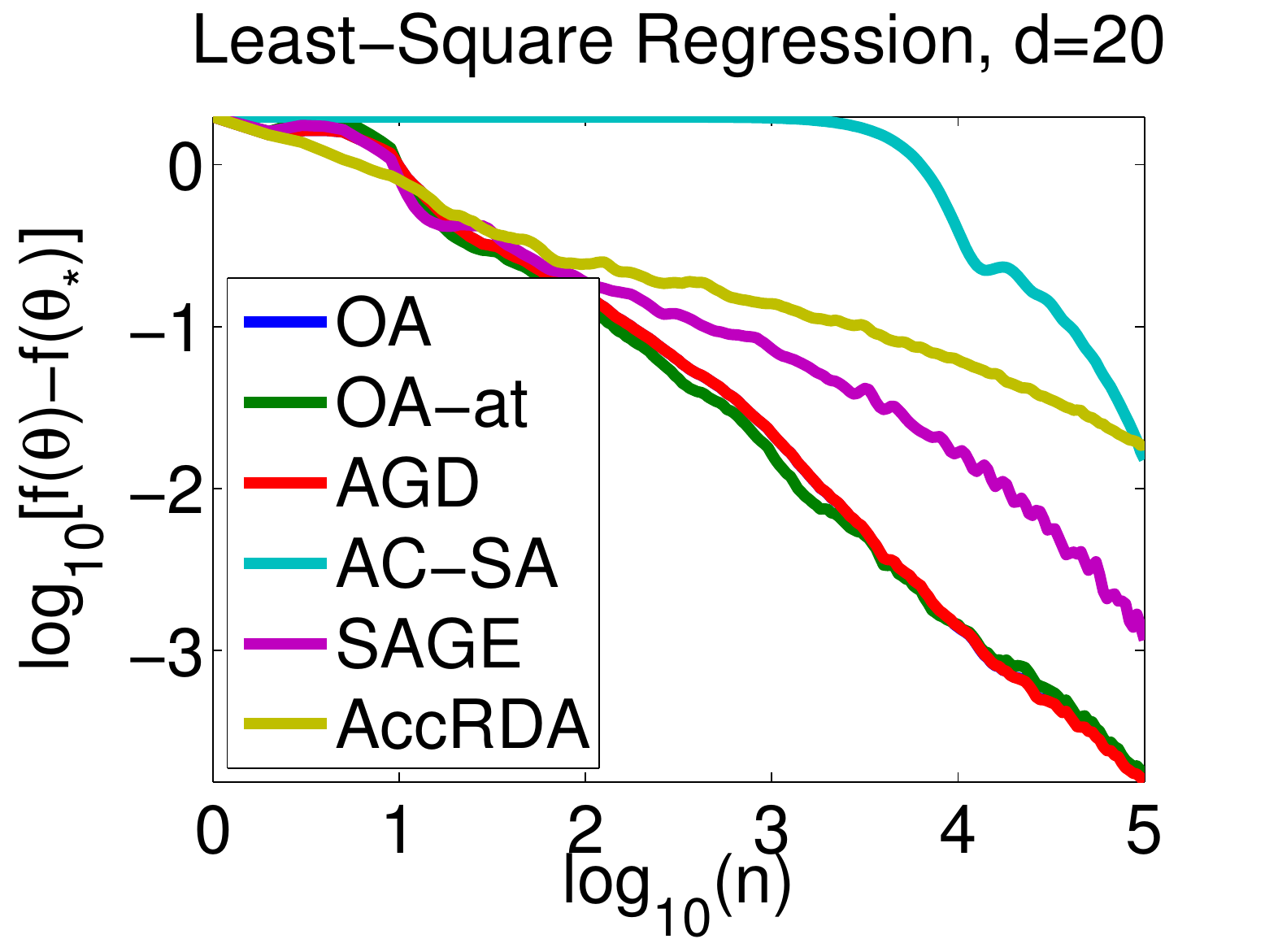}
   \end{minipage}
      \hspace*{.08\linewidth}
   \begin{minipage}[c]{.4 \linewidth}
\includegraphics[width=\linewidth]{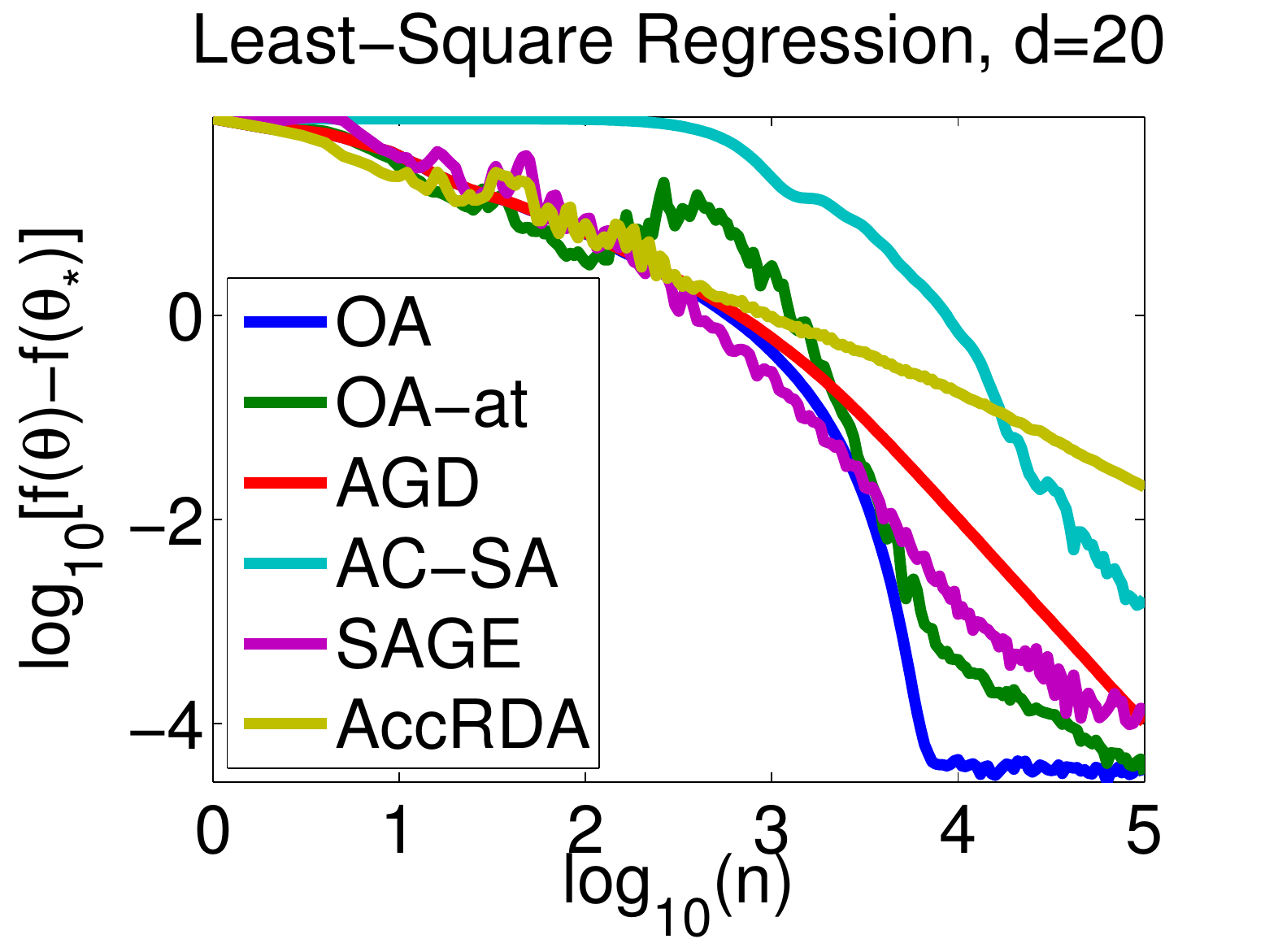}
      \end{minipage}

      \vspace*{-.3cm}

          \caption{Least-Square Regression. Left $\sigma=1$, $r=1$. Right $\sigma=0.1$, $r=10$.}

    \label{fig:plot7}
\end{figure}

\paragraph{Application to least-squares regression.}
We now see how these algorithms behave for least-squares regressions and the regular (non-homoscedastic) stochastic gradients described in \mysec{ls}.
We consider normally distributed inputs. The covariance matrix $H$ is the same as before. The outputs are generated from a linear function with homoscedatic noise with a signal-to-noise ratio of $\sigma$.
We consider $d=20$. We show   results averaged over 10 replications.
In Figure \ref{fig:plot7}, we consider again a situation where the bias dominates (left) and vice versa (right).
We see that our algorithm has the same good behavior than in the homoscedastic noise case and we conjecture that our bounds also hold in this situation.

\section{Conclusion}

\vspace*{-.1255cm}

We have provided a joint analysis of averaging and acceleration for non-strongly-convex quadratic functions in a single framework, both with noiseless and noisy gradients. This allows to define a class of algorithms that can benefit simultaneously of the known improvements of averaging and accelerations: faster forgetting of initial conditions (for acceleration), and better robustness to noise when the noise covariance is proportional to the Hessian (for averaging).

Our current analysis of our class of algorithms in \eq{thetafinal}, that considers two different affine combinations of previous iterates (instead of one for traditional acceleration), is limited to quadratic functions; an extension of its analysis to all smooth or self-concordant-like functions  would widen its applicability. Similarly, an extension to least-squares regression with natural heteroscedastic stochastic gradient, as suggested by our simulations, would be an interesting development.

\subsection*{Acknowledgements}
This work was partially supported by the MSR-Inria Joint Centre and a grant by the European Research Council (SIERRA project 239993).
 The authors would like to thank Aymeric Dieuleveut for helpful discussions.

 \bibliography{bio}
\bibliographystyle{plainnat}
\appendix

\section{Additional experimental results}
\label{app:det}

In this appendix, we provide additional experimental results to illustrate our theoretical results.

\subsection{Deterministic convergence}

\vspace*{-.1256cm}

\paragraph{Comparaison for $d=1$.}

\begin{figure}[!h]
\centering
\begin{minipage}[c]{.45\linewidth}
\includegraphics[width=\linewidth]{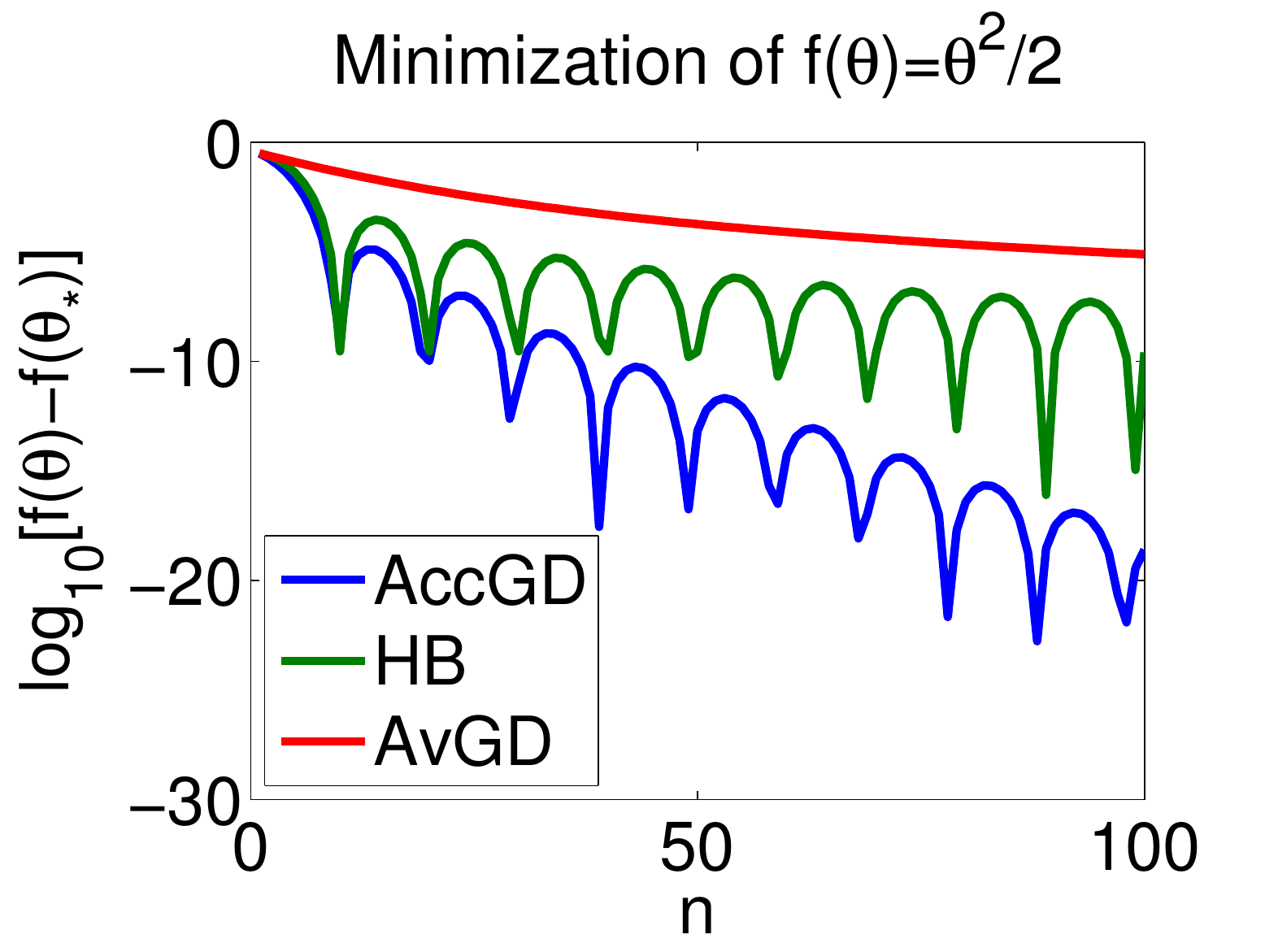}
   \end{minipage} \hspace*{.08\linewidth}
   \begin{minipage}[c]{.45\linewidth}
\includegraphics[width=\linewidth]{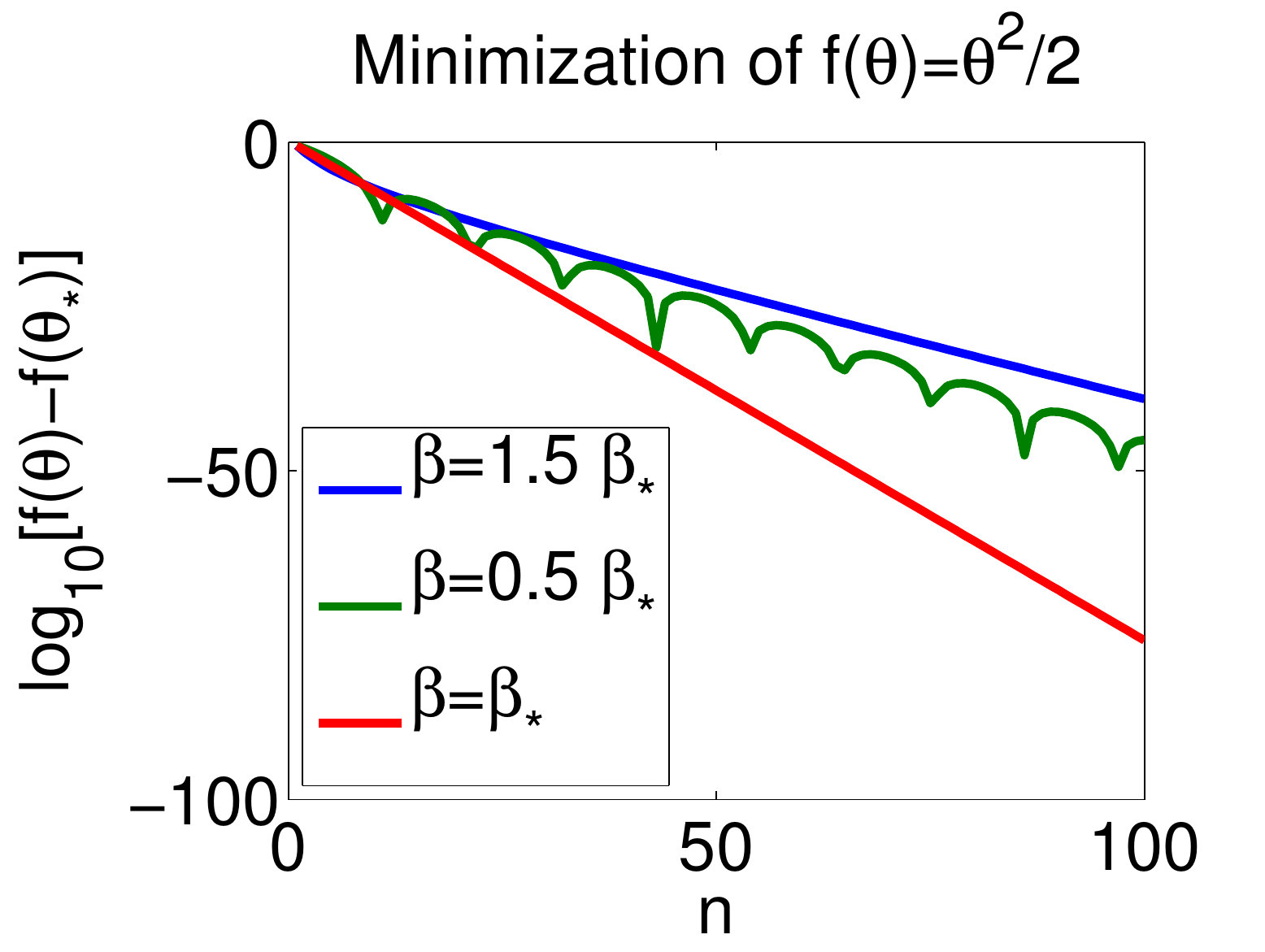}
   \end{minipage}

   \vspace*{-.5cm}

   \caption{Deterministic case for $d=1$ and $\alpha=1/10$. Left: classical algorithms, right: different oscillatory behaviors.}
   \label{fig:plot1}
\end{figure}

In Figure \ref{fig:plot1}, we minimize  a one-dimensional quadratic function $f(\theta)=\frac{1}{2}\theta^2$ for a fixed step-size $\alpha=1/10$ and different step-sizes $\beta$.
In the left plot, we compare Acc-GD, HB and Av-GD. We see that HB and Acc-GD both oscillate and that Acc-GD leverages strong convexity to converge faster.
In the right plot, we compare the behavior of the algorithm for different values of $\beta$.
We see that the optimal rate is achieved for $\beta=\beta_*$ defined to be the one for which there is a double coalescent eigenvalue, where the convergence is linear at speed $O(1-\sqrt{\alpha L})^n$.
When $\beta>\beta_*$, we are in the real case and when $\beta<\beta_*$ the algorithm oscillates to the solution.

\paragraph{Comparison between the different eigenspaces.}
\begin{figure}[!h]
\centering
\begin{minipage}[c]{.45\linewidth}
\includegraphics[width=\linewidth]{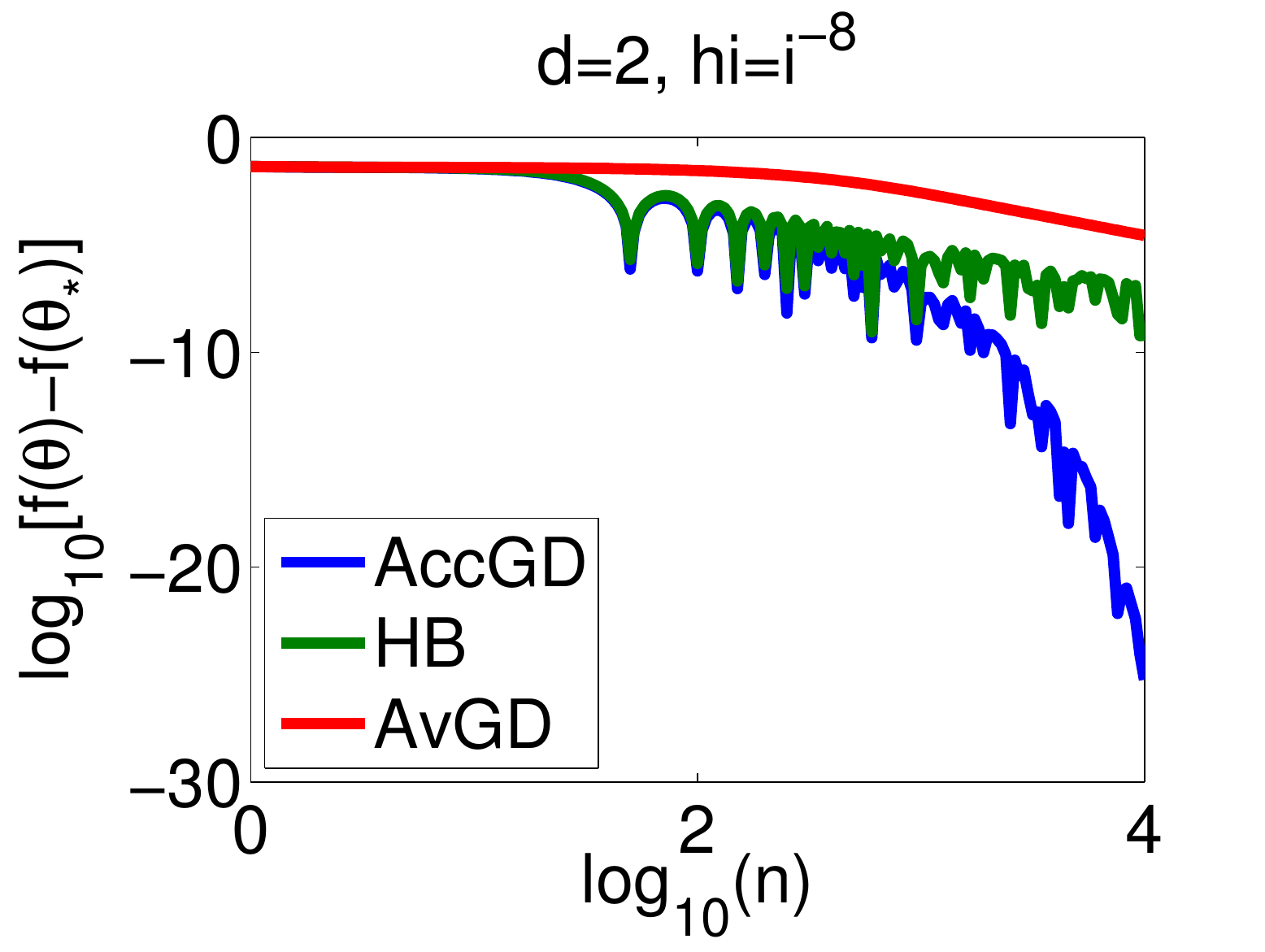}
   \end{minipage} \hspace*{.08\linewidth}
   \begin{minipage}[c]{.45\linewidth}
\includegraphics[width=\linewidth]{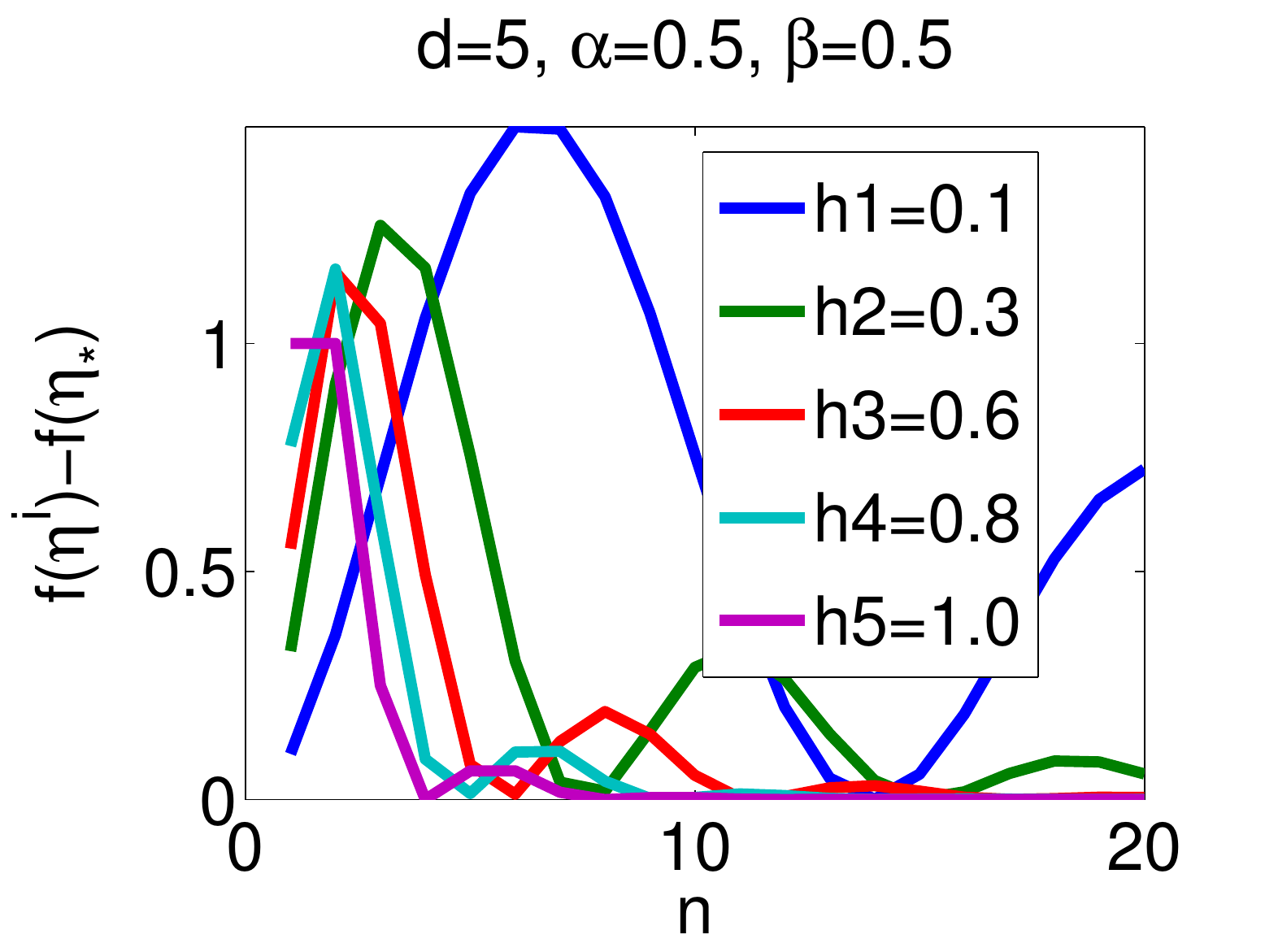}
   \end{minipage}

   \vspace*{-.25cm}

 \caption{Left: Deterministic quadratic optimization for $d=2$. Right: Function value of the projection of the iterate on the different eigenspaces ($d=5$).
   }
   \label{fig:plot2}
\end{figure}

Figure \ref{fig:plot2} shows interactions between different eigenspaces. In the left plot, we optimize a quadratic function of dimension $d=2$.
The first eigenvalue is $L=1$ and the second is $\mu=2^{-8}$.
For Av-GD the convergence is of order $O(1/n)$ since the problem is ``not'' strongly convex (i.e., not appearing as strongly convex since $n \mu $ remains small).
The convergence is at the beginning the same for HB and Acc-GD, with oscillation at speed $O(1/n^2)$, since the small eigenvalue prevents Acc-GD from having a linear convergence.
Then for large $n$, the convergence becomes linear for Acc-GD, since $\mu n$ becomes large.
In the right plot, we optimize a quadratic function in dimension $d=5$ with eigenvalues from $1$ to $0.1$.
We show the function values of the projections of the iterates $\eta_n$ on the different eigenspaces.
We see that high eigenvalues first dominate, but converge quickly to zero, whereas small ones keep oscillating, and converge more slowly.

\paragraph{Comparison for $d=20$.}
\begin{figure}[!h]
\centering
\begin{minipage}[c]{.45\linewidth}
\includegraphics[width=\linewidth]{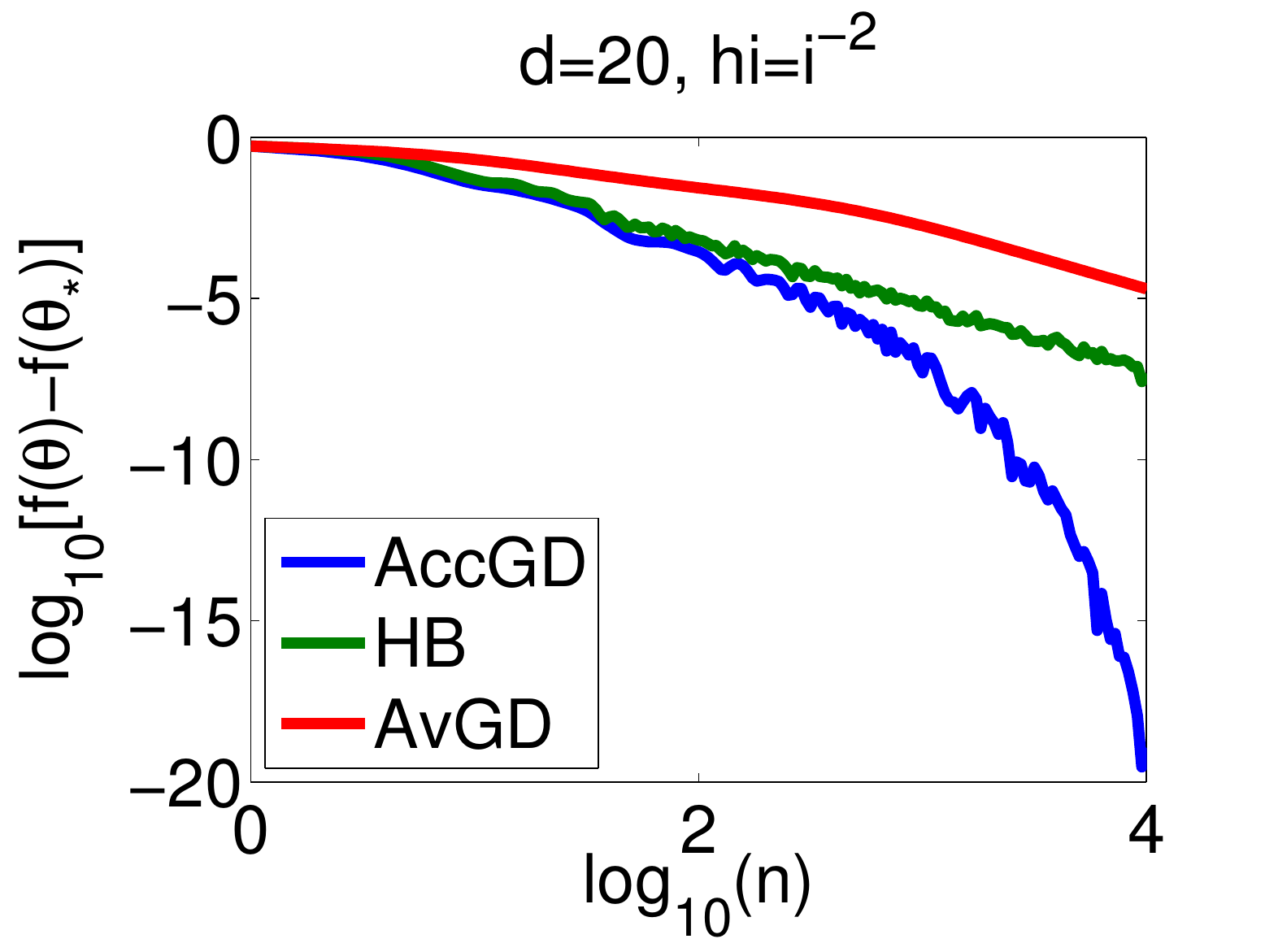}
   \end{minipage} \hspace*{.08\linewidth}
   \begin{minipage}[c]{.45\linewidth}
\includegraphics[width=\linewidth]{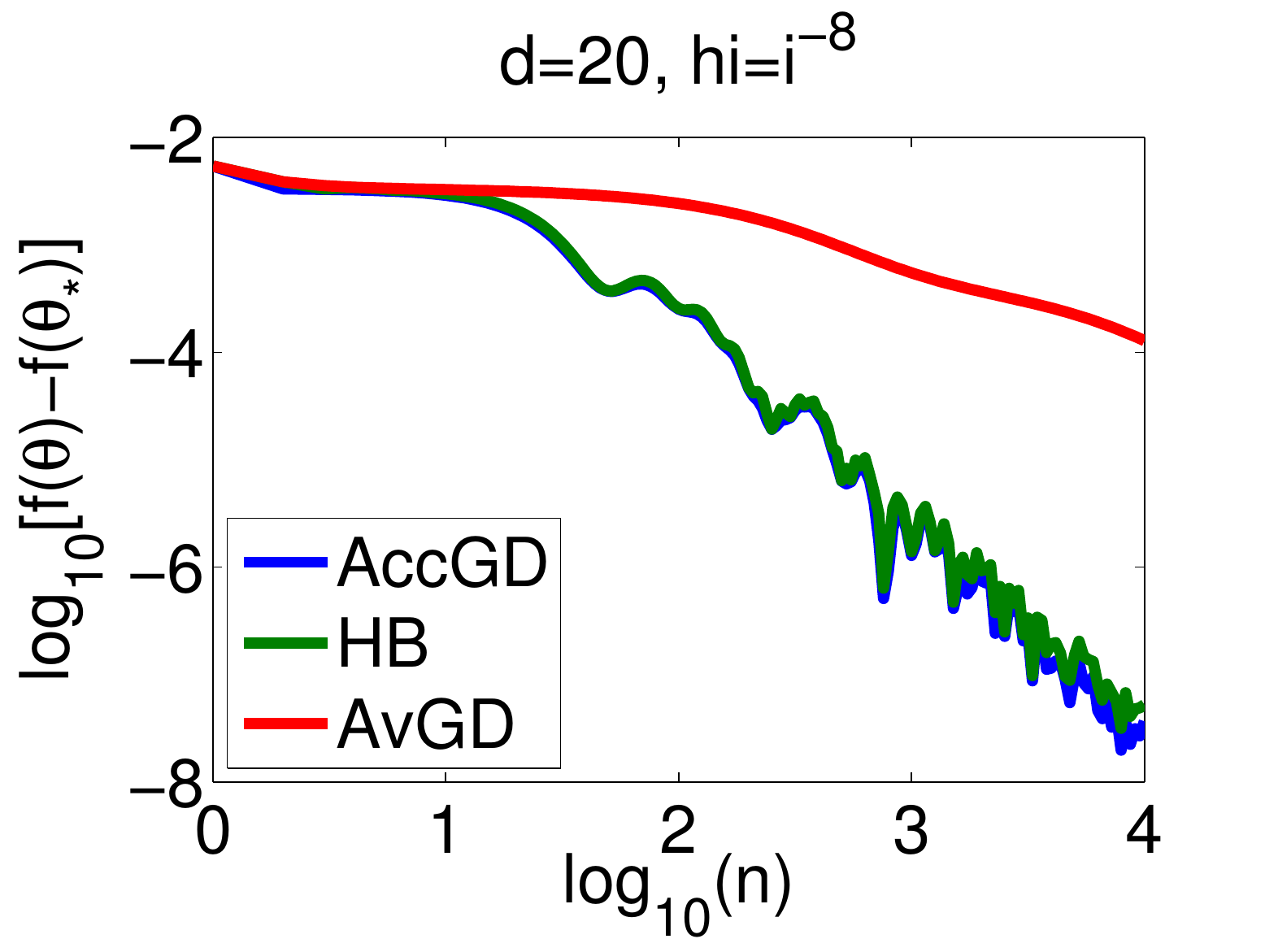}
   \end{minipage}

   \vspace*{-.25cm}

   \caption{Deterministic case for $d=20$ and $\gamma=1/10$. Left: $m=2$. Right: $m=8$.}
   \label{fig:plot3}
\end{figure}
In Figure \ref{fig:plot3}, we optimize two $20$-dimensional quadratic functions with different eigenvalues with Av-GD, HB and Acc-GD for a fixed step-size $\gamma=1/10$.
In the left plot, the eigenvalues are $1/k^2$ and in the right one, they are $1/k^8$, for $k=1,\dots,d$.
We see that in both cases, Av-GD converges at a rate of $O(1/n)$ and HB at a rate of $O(1/n^2)$.
For Acc-GD the convergence is linear when $\mu$ is large (left plot) and becomes sublinear at a rate of $O(1/n^2)$ when $\mu$ becomes small (right plot).

\subsection{Noisy convergence with unstructured additive noise}

\vspace*{-.1256cm}

We optimize the same quadratic function, but now with noisy gradients.
We compare our algorithm to other stochastic accelerated algorithms, that is,   AC-SA \citep{lan}, SAGE \citep{sage} and Acc-RDA \citep{xiao}, which are presented in Appendix~\ref{app:comparison}.
For all these algorithms (and ours) we take the optimal step-sizes defined in these papers.
We plot the results averaged over 10 replications.

\begin{figure}[!h]
\centering
\includegraphics[width=.6\linewidth]{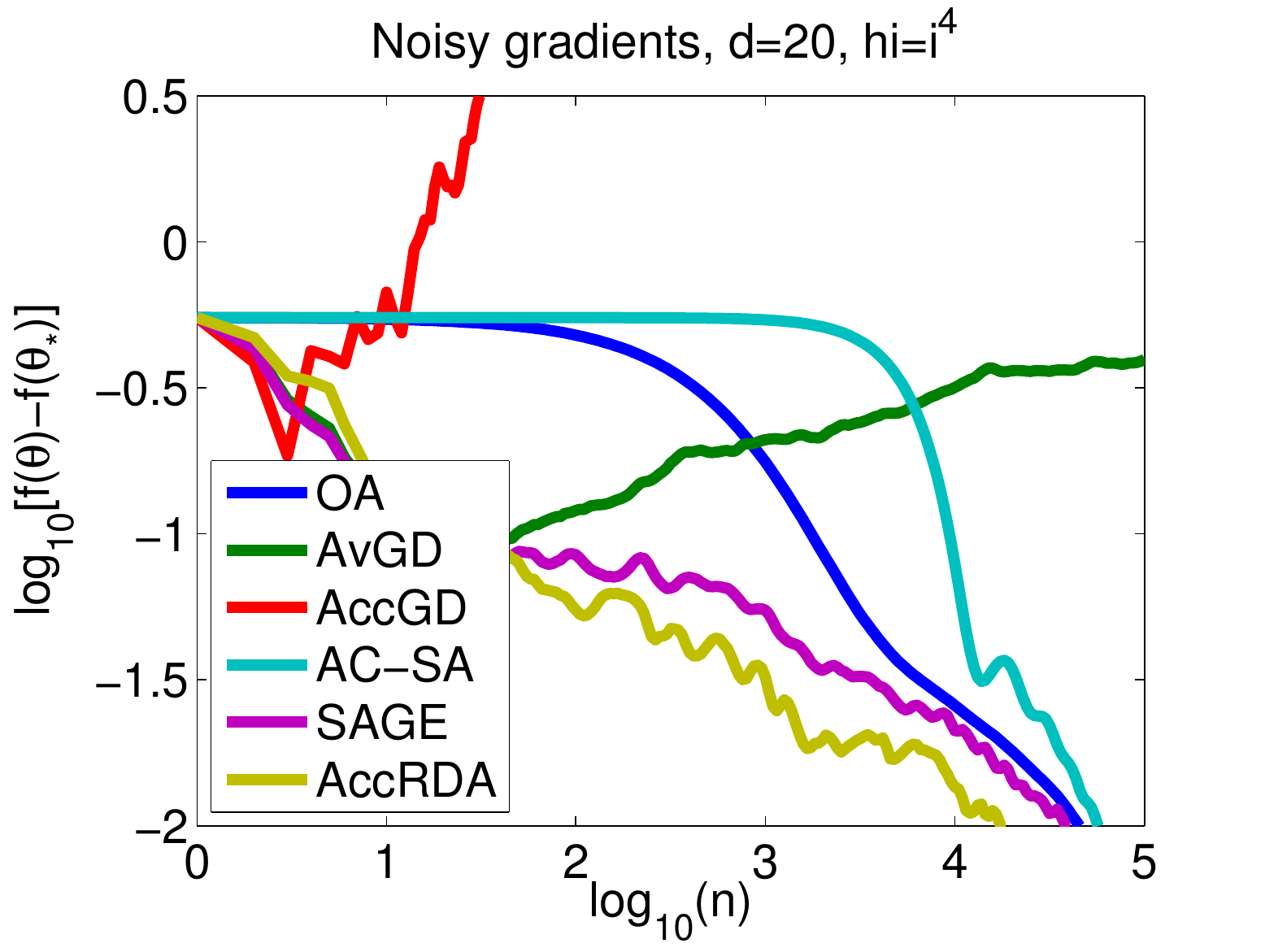}

\vspace*{-.25cm}

   \caption{Quadratic optimization with additive noise.}
   \label{fig:plot4}
\end{figure}

We   consider  in Figure \ref{fig:plot4} an i.i.d.~zero mean noise of variance $C=\idm$.
We see that all the accelerated algorithms achieve the same precision whereas Av-GD with constant step-size does not converge and Acc-Gd diverges.
However SAGE and AC-SA are anytime algorithms and are faster at the beginning since their step-sizes are decreasing and not a constant (with respect to $n$)  function of the horizon  $N$.

\section{Proofs of \mysec{soa}}
\label{app:soa}

\subsection{Proof of Theorem \ref{theo:1}}
\label{app:A1}
Let $(P_n,Q_n,R_n)\in(\RR[X])^3$ for all $n\in\NN$ be a sequence of polynomials. We consider the iterates defined for all $n\in\NN^*$ by 
\begin{equation*}
 \tnp=P_n(H)\tn+Q_n(H)\tnm+R(H)q,
\end{equation*}
started from $\theta_0=\theta_1\in \RR^d$.
The \ref{optimum-centering} property gives for $n\in\NN^*$:
 \begin{equation*}
  \theta_*=P_n(H)\theta_*+Q_n(H)\theta_*+R_n(H)q.
 \end{equation*}
Since $\theta_*=H^{-1}q$ we get for all  $q\in\RR^d$
 \begin{equation*}
 H^{-1}q=P_n(H)H^{-1}q+Q_n(H)H^{-1}q+R_n(H)q .
 \end{equation*}
For all $\tilde q \in \RR^d$ we apply this relation to vectors $q=H\tilde q$:
 \begin{equation*}
 \tilde q=P_n(H)\tilde q+Q_n(H)\tilde q+R_n(H)H\tilde q \text{\quad$\forall \tilde q\in\RR^d$},
 \end{equation*}
 and we get 
 \begin{equation*}
  I=P_n(H)+Q_n(H)+R_n(H)H \text{\quad $\forall n\in\NN^*$}.
 \end{equation*}
Therefore there are polynomials $(\bar P_n,\bar Q_n) \in(\RR[X])^2$ and $q_n\in\RR$ for all $n\in\NN^*$ such that we have for all $n\in\NN$: 
    \begin{eqnarray}
 P_n(X)&=&(1-q_n)I+X\bar P_n(X) \nonumber\\
 Q_n(X)&=&q_nI+X\bar Q_n(X) \nonumber\\
 R_n(X)&=&-(\bar P_n(X)+\bar Q_n(X)) \label{eq:poly}.
\end{eqnarray}
The  \ref{n-scalability} property means that there are polynomials $(P,Q)\in(\RR[X])^2$ independent of $n$ such that:
\BEAS
P_n(X)=\frac{n}{n+1}P(X),\\
Q_n(X)=\frac{n-1}{n+1}Q(X).
\EEAS
And in connection with \eq{poly} we can rewrite $P$ and $Q$ as:
\BEAS
P(X)=\bar p +X \bar P(X),\\
Q(X)=\bar q+X\bar Q(X),
\EEAS
with $(\bar p, \bar q)\in\RR^2$ and \mbox{$(\bar P, \bar Q)\in(\RR[X])^2$}. Thus for all $n\in\NN$:
    \begin{eqnarray}
q_n&=&\frac{n-1}{n+1}\bar q \label{eq:qbar}\\
\bar Q_n(X)&=&\frac{n-1}{n}Q(X)\nonumber \\
\frac{n}{n+1}\bar p&=&(1- q_n) \label{eq:pbar}\\
\bar P_n(X)&=&\frac{n}{n+1}P(X).\nonumber
    \end{eqnarray}
\eq{qbar} and \eq{pbar} give:
\begin{equation*}
 \frac{n}{n+1}\bar p=\left(1-\frac{n-1}{n+1}\bar q\right).
\end{equation*}
Thus for  $n=1$, we have  $\bar p =2$. 
Then \mbox{$-\frac{n-1}{n+1}\bar q =\frac{2n}{n+1}-1=\frac{n-1}{n+1}$} and $\bar q =-1$.
Therefore
\BEAS
 P_n(H)&=&\frac{2n}{n+1}I+\frac{n}{n+1}\bar P(H)H \\
 Q_n(H)&=&-\frac{n-1}{n}I+\bar Q(H)H \\
 R_n(H)&=&-\left(\frac{n \bar P (H)+(n-1)\bar Q(H)}{n+1}\right).
 \EEAS
 We let $\bar A=-(\bar P+\bar Q)$ and $\bar B=\bar Q$ so that we have:
 \BEAS
 P_n(H)&=&\frac{2n}{n+1}\left(I-\left(\frac{\bar A(H)+\bar B(H)}{2}\right)H\right) \\
 Q_n(H)&=&-\frac{n-1}{n}\left(I-\bar B(H)H\right) \\
 R_n(H)&=&\left(\frac{n \bar A (H)+\bar B(H)}{n+1}\right),
 \EEAS
 and with $\phi_n=\theta_n-\theta_*$ for all $n\in \NN$, the algorithm can be written under the form:
 \begin{equation*}
  \phi_{n+1}=\left[I-\left(\frac{n}{n+1}\bar A(H)+\frac{1}{n+1}\bar B(H) \right) H\right]\phi_n+\left(1-\frac{2}{n+1}\right)\left[I-\bar B(H) H\right](\phi_n-\phi_{n-1}).
 \end{equation*}

\subsection{Av-GD as two steps-algorithm}\label{sec:agdacc}
\label{app:A2}

We show now that when  the averaged iterate of Av-GD is seen as the main iterate we have that Av-GD with step-size $\gamma\in\RR$  is equivalent to:
$$ \tnp = \frac{2n}{n+1} \theta_n - \frac{n-1}{n+1} \theta_{n-1}-\frac{\gamma}{n+1} f' \big(
n\theta_n - (n-1)\theta_{n-1} 
\big).
 $$
We remind  
\BEAS
  \psi_{n+1}&=& \psi_n -\gamma f'(\psi_n),\\
  \theta_{n+1}&=&\theta_n+\frac{1}{n+1}(\psi_{n+1}-\tn).
\EEAS
Thus, we have:
    \begin{eqnarray}
  \tnp&=&\tn+\frac{1}{n+1}(\psi_{n+1}-\tn) \nonumber \\
   &=&\tn+\frac{1}{n+1}( \psi_{n}-\gamma f'(\psi_n)-\tn) \nonumber \\
&=&\tn+\frac{1}{n+1}( \tn+(n-1)(\tn-\tnm) -\gamma f'(\tn+(n-1)(\tn-\tnm))-\tn) \nonumber \\
&=& \frac{2n}{n+1} \theta_n - \frac{n-1}{n+1} \theta_{n-1} -\frac{\gamma}{n+1} f'(n\tn-(n-1)\tnm). \nonumber 
\end{eqnarray}

\section{Proof of \mysec{conv}}
\label{app:appB}

\subsection{Proof of  Lemma \ref{lemma:stabreal}}
 \begin{figure}[!h]
\centering
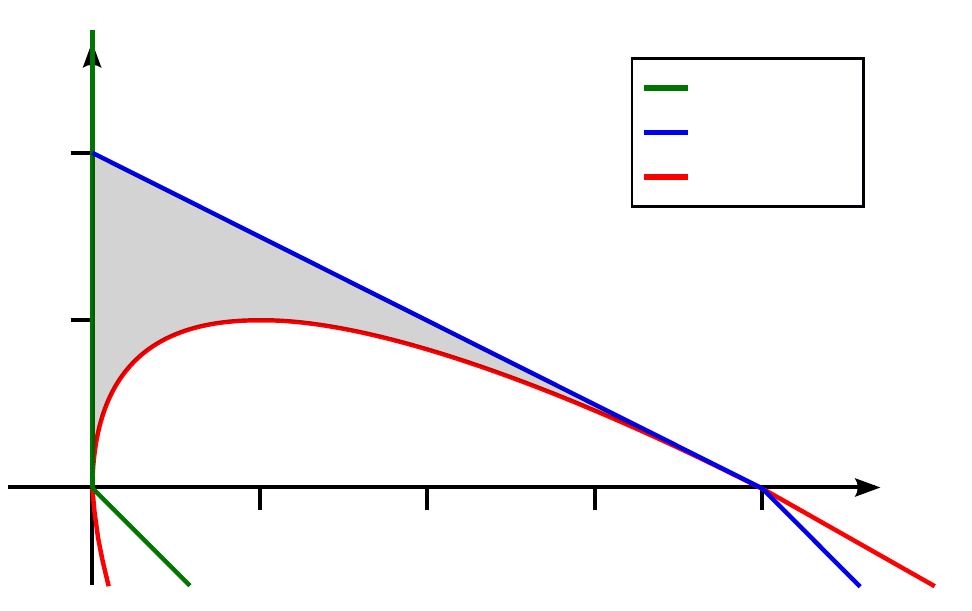
\caption{Stability in the real case, with all contraints plotted.}\label{fig:stabreal}
\end{figure}

The discriminant $\Delta_i$ is strictly positive when  $\big(\frac{\alpha+\beta}{2}\big)^2 h_i-\al > 0 $.
This is always true for $\alpha$ strictly negative.
For $\alpha$ positive and for $h_i\neq0$, this is true for $\vert \frac{\alpha+\beta}{2}\vert> \sqrt {\al/h_i}$ .
Thus the discriminant $\Delta_i$ is strictly positive for
\begin{eqnarray}
 \alpha< 0 &&\text{\quad or\quad}\nonumber \\
 \alpha\geq0&& \text{\quad and \quad } \left\{\beta < -\alpha-2\sqrt{\alpha/h_i} \text{\quad or\quad} \beta> -\alpha+2\sqrt{\alpha/h_i}\right\}.\nonumber
\end{eqnarray}
Then we determine when the modulus of the eigenvalues is less than one (which corresponds to \mbox{$-1\leq r_i^-\leq r_i^+\leq1$}).
\begin{eqnarray}
 r_i^+\leq 1 &\Leftrightarrow& \sqrt{ h_i\left(\left(\frac{\al +\be}{2} \right)^2h_i-\al \right)}\leq 	\left(\frac{\be+\al}{2}\right) h_i \nonumber \\
 &\Leftrightarrow & h_i\left(\left(\frac{\be+\al}{2} \right)^2h_i-\al \right) \leq \left[\left(\frac{\be+\al}{2}\right)h_i\right]^2 \text{ and\quad } \frac{\al+\be}{2} \geq 0\nonumber \\
 &\Leftrightarrow & h_i \al \geq 0 \text{ \quad and \quad } \frac{\al+\be}{2} \geq 0\nonumber \\
  &\Leftrightarrow & \al \geq 0 \text{ \quad and \quad  } \al+\beta \geq 0. \nonumber
\end{eqnarray}
Moreover, we have :
\begin{eqnarray}
 \!\!\! r_i^-\geq -1 &\Leftrightarrow& \sqrt{ h_i\left(\left(\frac{\be+\al}{2} \right)^2h_i-\al \right)}\leq 	2-\left(\frac{\be+\al}{2}\right) h_i \nonumber \\
  &\Leftrightarrow& { h_i\left(\left(\frac{\be+\al}{2} \right)^2h_i-\al \right)}\leq 	\left[2-\left(\frac{\be+\al}{2}\right) h_i \right]^2\text{ and } 2-\left(\frac{\be+\al}{2}\right) h_i \geq 0\nonumber \\
    &\Leftrightarrow& { h_i\left(\left(\frac{\be+\al}{2} \right)^2h_i-\al \right)}\leq 	4-4\left(\frac{\be+\al}{2}\right) h_i+\left[\left(\frac{\be+\al}{2}\right) h_i  \right]^2\text{ and } \left(\frac{\be+\al}{2}\right)  \leq 2/h_i\nonumber \\
        &\Leftrightarrow& -h_i\al \leq 	4-4\left(\frac{\be+\al}{2}\right) h_i\text{ and } \frac{\be}{2}  \leq 2/h_i-\frac{\al}{2}\nonumber \\
        &\Leftrightarrow& \be  \leq 	2/h_i-\al/2 \text{ and } \be  \leq 4/h_i-\al. \nonumber
\end{eqnarray}
Figure \ref{fig:stabreal} (where we plot all the constraints we have so far) enables to conclude that the discriminant $\Delta_i$ is strictly positive  and the algorithm is stable when the following three conditions are satisfied:
\begin{eqnarray}
 \alpha&\geq&0 \nonumber \\
 \alpha+2\beta&\leq& 4/h_i \nonumber \\
 \alpha+\beta&\geq& 2\sqrt{\alpha/h_1}. \nonumber
\end{eqnarray}
For any of those $\alpha$ et $\beta$ we will have:
\begin{equation*}
 \eta_n^i=c_1{(r_i^-)}^n+c_2{(r_i^+)}^n.
\end{equation*}
Since $\eta_0^i=0$, $c_1+c_2=0$ and for $n=1$, $c_1=\eta^i_1/(r_i^--r_i^+)$; we thus have:
\begin{equation*}
 \eta_n^i=\frac{\eta_1^i}{2}\frac{{(r_i^+)}^n-{(r_i^-)}^n}{\sqrt{\Delta_i}}.
\end{equation*}
Thus, we get the final expression:
\begin{equation*}
 ({\phi_n^i})^2h_i=\frac{({\phi_1^i})^2}{4n^2}\frac{\left\{\left[r_i+\sqrt{\Delta_i}\right]^n-\left[r_i-\sqrt{\Delta_i}\right]^n\right\}^2}{\Delta_{i}/h_{i}}.
\end{equation*}

\subsection{Proof of Lemma \ref{lemma:stabcom}}

 \begin{figure}[!h]
\centering
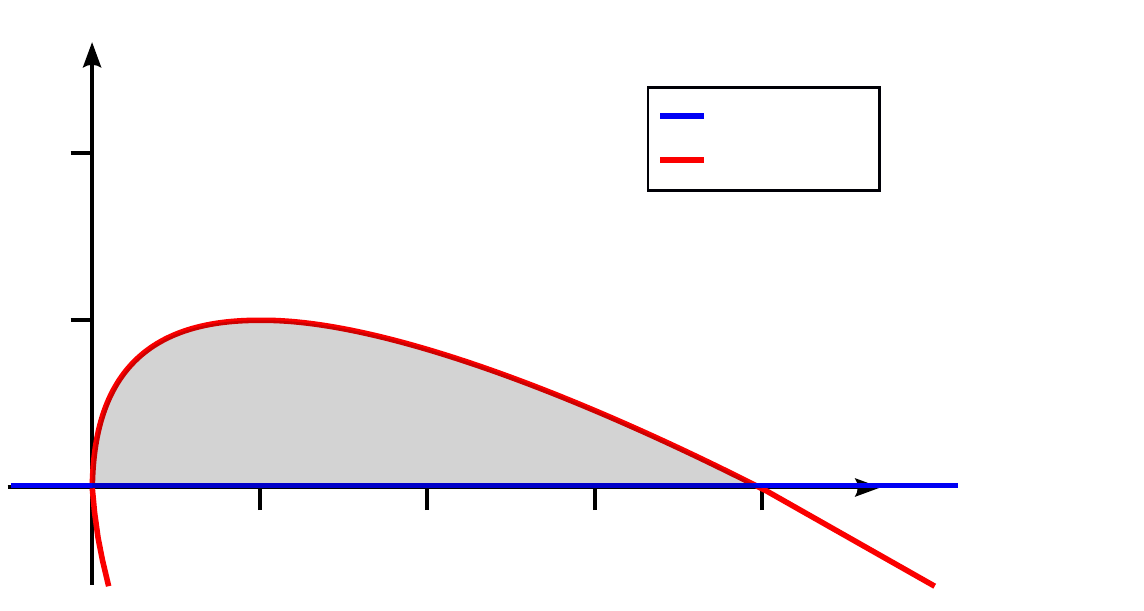
\caption{Stability in the complex case, with all constraints plotted.}\label{fig:stabcomp}
\label{fig:complex}
\end{figure}

The discriminant $\Delta_i$ is strictly negative if and only if  $\big(\frac{\alpha+\be}{2}\big)^2 h_i-\al< 0 $.
This implies $\vert \frac{\alpha+\be}{2}\vert< \sqrt{\al/h_i}$.
The modulus of the eigenvalues is $\vert r_i^\pm\vert^2 = 1-\be h_i$.
Thus the discriminant $\Delta_i$ is  strictly negative and  the algorithm is stable for
 \begin{eqnarray}
 \alpha,\beta&\geq &0 \nonumber \\
\alpha+\beta&< &\sqrt{\alpha/h_i}, \nonumber
\end{eqnarray}
as shown in Figure \ref{fig:complex}.

For any of those $\alpha$ et $\beta$ we have:
\begin{equation*}
 \eta_n^i=[c_1\cos(\omega_i n)+c_2\sin(\omega_i n)]\rho_i^n,
\end{equation*}
with $\rho_i=\sqrt{ 1-\beta h_i  }$, $\sin (\omega_i)=\sqrt{-\Delta_i}/\rho_i$ and $\cos(\omega_i)=r_i/\rho_i$.
Since $\eta_0^i=0$, $c_1=0$ and we have for $n=1$, $c_2=\eta^i_1/(\sin(\omega_i )\rho_i)$. Therefore

\begin{equation*}
 \eta_n^i=\eta_1^i\frac{\sin(\omega_i n)}{\sqrt{ -\Delta_i}}(1-\beta h_i)^{n/2},
\end{equation*}
and
\begin{equation*}
 ({\phi_n^i})^2h_i=\frac{({\phi_1^i})^2}{n^2}\frac{\sin^2(\omega_i n)}{\sin^2(\omega_i )/h_{i}}(1-\beta h_i)^{n-1}.
 \end{equation*}

\subsection{Coalescing eigenvalues}\label{app:coal}

When $\beta=2\sqrt{\alpha/h_i}-\alpha$, the discriminant $\Delta_{i}$ is equal to zero and we have a double real eigenvalue:
 \begin{equation*}
  r_i=1-\sqrt{\alpha h_i}.
 \end{equation*}
Thus the algorithm is stable for $\al < \frac{4}{h_i}$. For any of those $\alpha$ et $\beta$ we have:
 \begin{equation*}
  \eta_n^i=(c_1+nc_2)r^n.
 \end{equation*}
 This gives with  $\eta_0^i=0$, $c_1=0$ and $c_2=\eta_1^i/r$.
 Therefore
\begin{equation*}
 \eta_n^i=n\eta_i(1-\sqrt{\alpha h_i})^{n-1},
\end{equation*}
and:
\begin{equation*}
 ({\phi_n^i})^2h_i=h_i({\phi_1^i})^2(1-\sqrt{\alpha h_i})^{2(n-1)}.
\end{equation*}

In the presence of coalescing eigenvalues the convergence is linear if $0<\alpha<4/h_i$ and $h_i>0$,
however one might worry about the behavior of $(({\phi_n^i})^2h_i)_n$ when $h_i$ becomes small.
Using the bound $x^2\exp(-x)\leq 1$ for $x\leq1$, we have for $\alpha<4/h_{i}$:
\begin{eqnarray}
 h_i (1-\sqrt{\alpha h_i})^{2n}&=& h_i\exp(2n\log(\vert 1-\sqrt{\alpha h_i}\vert))\nonumber \\
 &\leq& hi\exp(-2n \min\{\sqrt{\alpha h_i},2-\sqrt{\alpha h_i}\}) \nonumber \\
  &\leq& \frac{h_i}{\min\{\sqrt{\alpha h_i},2-\sqrt{\alpha h_i}\}^2} \nonumber \\
  &\leq&\max\left\{\frac{1}{\alpha},\frac{h_i}{(2-\sqrt{\alpha h_i})^2} \right\}. \nonumber
\end{eqnarray}
Therefore we always have the following bound for $\alpha< 4/h_i$:
\begin{equation*}\label{lemma:boundcoa}
 ({\phi_n^i})^2h_i\leq \frac{({\phi_1^i})^2} {4n^{2}}\max \left\{ \frac{1}{ \alpha}  , \frac{h_{i}}{(2-\sqrt{\alpha h_{i}})^{2}} \right\}.
\end{equation*}
Thus for $\alpha h_i \leq 1$ we get:
\begin{equation*}
  ({\phi_n^i})^2h_i\leq \frac{({\phi_1^i})^2} {4n^{2}\alpha}.
\end{equation*}

\section{Proof of Theorem \ref{theo:bounddet}}\label{app:bounddet}

\subsection{Sketch of the proof}
 \begin{figure}[!h]
\centering
\begin{minipage}[c]{.4\linewidth}
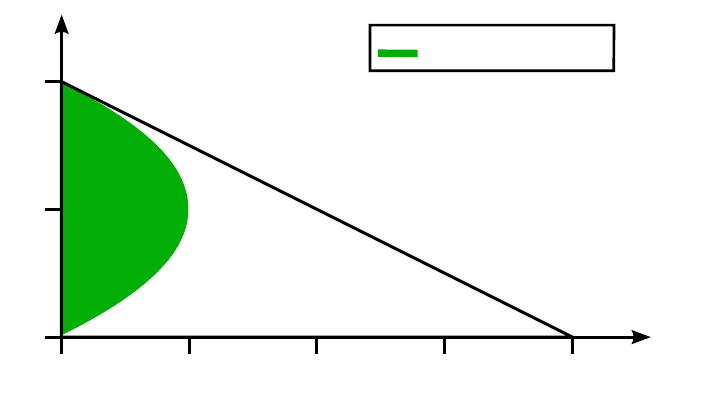
\vspace*{-0.8cm}
\caption{Validity of Lemma \ref{lemma:l1}}
\label{fig:l1}
\vspace*{+.75cm}

   \end{minipage}
   \hspace*{.08\linewidth}
   \begin{minipage}[c]{.4\linewidth}
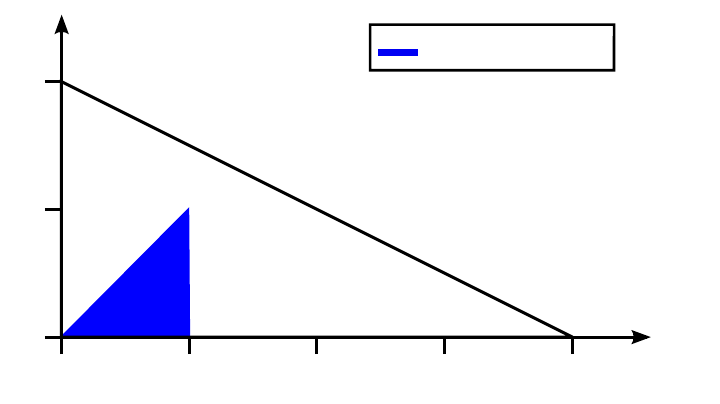
\vspace*{-0.8cm}
\caption{Validity of Lemma \ref{lemma:l2}}
\label{fig:l2}
\vspace*{+.75cm}
   \end{minipage}

    \begin{minipage}[c]{.4\linewidth}
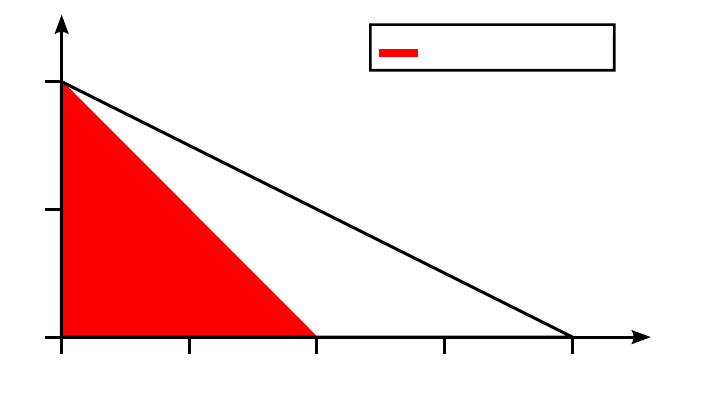
\vspace*{-0.8cm}
\caption{Validity of Lemma \ref{lemma:l3}}
\label{fig:l3}
   \end{minipage}
    \hspace*{.08\linewidth}
 \begin{minipage}[c]{.4\linewidth}
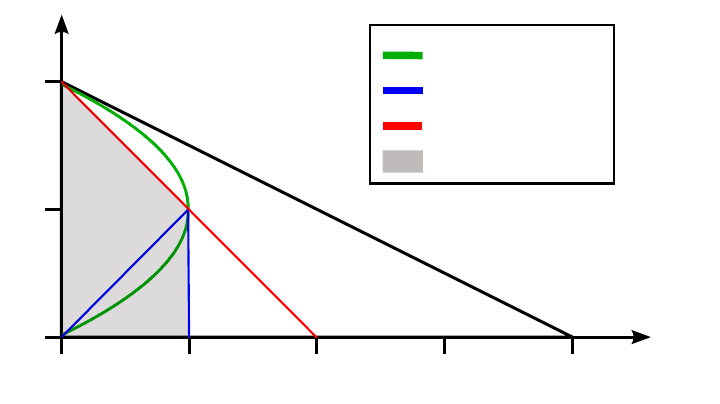
\vspace*{-0.8cm}
\caption{Area of Theorem \ref{theo:bounddet}}
\label{fig:theo}
   \end{minipage}
\end{figure}

 We divide the domain of validity of Theorem \ref{theo:bounddet} in three subdomains as explained in Figure \ref{fig:theo}.
 On the domain described in Figure \ref{fig:l1} we have a first bound on the iterate $\eta_n^i$:
\begin{lemma}\label{lemma:l1}
 For $ 0\leq \alpha\leq 1/h_i$ and  $ 1-\sqrt{1-\alpha h_i}<\beta h_i <1+\sqrt{1-\al h_i }$, we have:
\begin{equation*}
 (\eta_n^i)^2\leq\frac{(\eta^i_1)^2}{\alpha h_i}.
\end{equation*}
\end{lemma}
And on the domain described Figure \ref{fig:l2} we also have:
\begin{lemma}\label{lemma:l2}
For $0\leq\alpha\leq 1/h_i$ and $\beta\leq \alpha$ we have:
\begin{equation*}
 (\eta_n^i)^2\leq \frac{2(\eta^i_1)^2}{\alpha h_i}.
\end{equation*}
\end{lemma}
These two lemmas enable us to prove the first bound of Theorem \ref{theo:bounddet} since the domain of this theorem is included in the intersection of the two domains of these lemmas as shown in Figure \ref{fig:theo}.

Then we have the following bound on domain described in Figure \ref{fig:l3}:

\begin{lemma}\label{lemma:l3}
 For $0\leq\alpha\leq2/h_i$ and $0\leq\beta\leq2/h_i-\alpha$, we have:
 \begin{equation*}
          \vert \eta_n^i \vert \leq \min\left\{\frac{2 \sqrt{2n}}{\sqrt{(\alpha+\beta)h_i}}, \frac{4}{(\alpha+\beta)h_i}\right\}.
         \end{equation*}
\end{lemma}

Since the domain of definition of Theorem \ref{theo:bounddet} is included in the domain of definition of Lemma \ref{lemma:l3} (as shown in Figure \ref{fig:theo}), this lemma proves the last two bounds of the theorem.

\subsection{Outline of the proofs of the Lemmas}
\BIT
\item[--]
We find a Lyapunov function $G$ from $\RR^2$ to $\RR$ such that the sequence $(G(\eta^i_n,\eta^i_{n-1}))$ decrease along the iterates.
\item[--]
We also prove that $G(\eta^i_n,\eta^i_{n-1})$ dominates $c\Vert \eta^i_n\Vert^2$ when we want to have a bound on $\Vert \eta^i_n\Vert^2$ of the form \mbox{$\frac{1}{c}G(\eta^i_1,\eta^i_{0})=\frac{1}{c}G(\theta_0^i-\theta_*^i,0)$}.
\EIT

For readability, we remove the index $i$ and take $h_i=1$ without loss of generality.

\subsection{Proof of Lemma \ref{lemma:l1}}

We first consider a quadratic Lyapunov function $\begin{pmatrix} \eta_n \\ \eta_{n-1} \end{pmatrix}^{\top} G_1\begin{pmatrix} \eta_n \\ \eta_{n-1} \end{pmatrix}$ with $G_1=\begin{pmatrix}1 & \al -1\\ \al -1 & 1-\al   \end{pmatrix}$.
We note that $G_1$ is symmetric positive semi-definite for $\alpha\leq1$.
We recall $F_{i}=\begin{pmatrix}
                 2-(\alpha+\beta)&\beta -1 \\
                1 &0
                \end{pmatrix}$.

For the result to be true we need  for $ 0\leq \alpha\leq 1$ and  $ 1-\sqrt{1-\alpha}<\beta <1+\sqrt{1-\al }$ \ two properties:
 \begin{equation}\label{eq:PP}
  F_i^{\top}G_1F_i \preccurlyeq G_1,
 \end{equation}
and
\begin{equation}\label{eq:p1}
 \alpha \begin{pmatrix}1 &0\\ 0 &0   \end{pmatrix} \preccurlyeq G_1.
\end{equation}

\paragraph{Proof of \eq{p1}.}
 We have:
\begin{equation*}
G_1-\alpha  \begin{pmatrix}1 &0\\ 0 &0   \end{pmatrix}=(1-\alpha )\begin{pmatrix}1 &1\\ 1 &1   \end{pmatrix}\ \succcurlyeq 0 \text{\quad for $\alpha\leq 1$}.
\end{equation*}

\paragraph{Proof of \eq{PP}.}

Since $\beta\mapsto  F_i(\beta)^{\top}G_1F_i(\beta)-G_1$ is convex in $\beta$ ($G_1$ is symmetric positive semi-definite), we only have to show \eq{PP} for the boundaries of the interval in $\beta$.
For $x\in\RR^*_+$:
  \begin{equation*}
   \begin{pmatrix}x^2-x& x\\ 1 & 0 \end{pmatrix}^{\top}\begin{pmatrix}1& -x^2\\ -x^2 & x^2 \end{pmatrix}\begin{pmatrix}x^2-x& x\\ 1 & 0 \end{pmatrix}-\begin{pmatrix}1& -x^2\\ -x^2 & x^2 \end{pmatrix}
   =-(1-x^2)^2\begin{pmatrix}
                            1&0\\0 & 0\end{pmatrix}\preccurlyeq0.
  \end{equation*}
This especially shows \eq{PP} for the boundaries of the interval with $x=\pm\sqrt{1-\alpha}$.

\paragraph{Bound.}
Thus, because $\eta_0=0$, we have
\begin{equation*}
      \alpha \eta_{n+1}^2\leq \Theta_n^{\top} G_1 \Theta_n \leq \Theta_{n-1}^{\top} G_1 \Theta_{n-1}\leq  \Theta_{0}^{\top} G_1 \Theta_{0}\leq \eta_1^2.
     \end{equation*}
This shows that for $ 0\leq \alpha\leq 1/h_i$ and  $ 1-\sqrt{1-\alpha h_i}<\beta h_i <1+\sqrt{1-\al h_i }$:
\begin{equation*}
 (\eta_n^i)^2\leq\frac{(\eta^i_1)^2}{\alpha h_i}.
\end{equation*}

\subsection{Proof of Lemma \ref{lemma:l2}}
We consider now a second Lyapunov function $G_2(\eta_n,\eta_{n-1})=(\eta_n-r\eta_{n-1})^2-\Delta (\eta_{n-1})^2$.
We have:
\begin{eqnarray}
G_2(\eta_n,\eta_{n-1})&=& (\eta_n-r \eta_{n-1})^2-\Delta\eta_{n-1}^2\nonumber \\
&=&(r \eta_{n-1}-(1-\beta) \eta_{n-2})^2-\Delta\eta_{n-1}^2\nonumber \\
 &=& (r^2-\Delta)\eta_{n-1}^2+(1-\beta)^2\eta_{n-2}^2-2(1-\beta) r \eta_{n-1}\eta_{n-2} \nonumber \\
 &=& ((1-\beta)\eta_{n-1}^2+(1-\beta)(r^2-\Delta)\eta_{n-2}^2-2(1-\beta) r \eta_{n-1}\eta_{n-2} \nonumber \\
 &=& (1-\beta)[(\eta_{n-1}-r\eta_{n-2})^2-\Delta(\eta_{n-2})^2].  \nonumber \\
 &=&(1-\beta)G_2(\eta_{n-1},\eta_{n-2}). \nonumber
\end{eqnarray}
Where we have used twice $r^2-\Delta=(1-\beta)$ and $\eta_n=2r\eta_{n-1}-(1-\beta)\eta_{n-2}$.
Moreover $G_2(\eta_n,\eta_{n-1})$ can be rewritten as:
\begin{equation*}
 G_2(\eta_n,\eta_{n-1})=(1-\frac{\alpha+\beta}{2})(\eta_n-\eta_{n-1})^2+\frac{\alpha-\beta}{2}(\eta_{n-1})^2+\frac{\alpha+\beta}{2}(\eta_{n})^2.
\end{equation*}
Thus for $\alpha+\beta\leq 2$ and $\beta\leq \alpha$ we have:
\begin{equation*}
 \frac{\alpha}{2} (\eta_{n})^2\leq G_2(\eta_n,\eta_{n-1}) =(1-\beta )^{n-1} G_2(\eta_1,\eta_{0})=(1-\beta )^{n-1}(\eta_1)^2.
\end{equation*}
Therefore for $\alpha+\beta\leq 2/h_i$ and $\beta\leq \alpha$, we have:
\begin{equation*}
 (\eta_n^i)^2\leq \frac{2(\eta_1^i)^2}{\alpha h_i}.
\end{equation*}

\subsection{Proof of Lemma \ref{lemma:l3}}

We may write $\eta_n$ as
\begin{equation*}
 \eta_n=r\eta_{n-1}+(r_+)^n+(r_-)^n.
\end{equation*}
Moreover, we have:
\begin{equation*}
 \vert(r_+)^n+(r_-)^n\vert\leq 2,
\end{equation*}
therefore for $\alpha+\beta\leq 2$,
\begin{equation*}
 \vert\eta_n\vert\leq r  \vert \eta_{n-1}\vert +2\leq 2 \frac{1-r^n}{1-r}\leq 2\frac{1-(1-(\frac{\alpha+\beta}{2}))^n}{(\frac{\alpha + \beta}{2})}.
\end{equation*}
Thus
\begin{equation*}
 \vert \eta_n\vert \leq \frac{2}{(\frac{\alpha+\beta}{2})h}.
\end{equation*}
Moreover for all $u\in[0,1]$ and $n\geq1$ we have $1-(1-u)^n\leq\sqrt{nu}$, since $1-(1-u)^n\leq1$ and $1-(1-u)^n=u \sum (1-u)^k \leq n u $.
Thus \begin{equation*}
\vert \eta_n\vert \leq \frac{2 \sqrt{n}}{\sqrt{(\frac{\alpha +\beta}{2})}}.
     \end{equation*}

Therefore for $0\leq\alpha\leq2/h_i$ and $\alpha+\beta \leq 2/h_i$ we have:
     \begin{equation*}
          \vert \eta^i_n \vert \leq \min\left\{\frac{2 \sqrt{2n}}{\sqrt{(\alpha+\beta)h_i}}, \frac{4}{(\alpha+\beta)h_i}\right\}.
         \end{equation*}

 \section{Lower bound}\label{app:lb}

 We  have the following lower-bound for the bound shown in Corollary~\ref{cor:det}, which shows that depending on which of the two terms dominates, we may always find a sequence of functions that makes it tight.

 \begin{proposition}\label{prop:ub}
 Let $L\geq0$.  For all sequences $0\leq \alpha_{n}\leq 1/L$ and $0\leq \beta_{n}\leq 2/L-\alpha_{n}$, such that $\alpha_{n}+\beta_{n}=o(n\alpha_{n})$ there exists a sequence of one-dimensional quadratic functions $(f_n)_{n}$ with second-derivative less than $   L $ such that:
 \begin{equation*}
 \lim \alpha_{n}n^2(f_n(\theta_n)-f_n(\theta_*))=  \frac{\Vert \theta_0-\theta_*\Vert^2}{2}.
 \end{equation*}
 For all sequences $0\leq \alpha_n\leq 1/L$  and $0\leq \beta_{n} \leq 2/L-\alpha_n$, such that ${n\alpha_{n}}=o({\alpha_{n}+\beta_{n}})$, there exists a sequence of one-dimensional quadratic functions $(g_n)_{n}$ with second-derivative less than $   L $ such that:
 \begin{equation*}
 \lim n(\alpha_n+\beta_{n})(g_n(\theta_n)-g_n(\theta_*))= \frac{\left(1-\exp(-2)\right)^2\Vert \theta_0-\theta_*\Vert^2}{4}.
 \end{equation*}
 \end{proposition}

 \paragraph{Proof of the first lower-bound.}

 For the first lower bound we consider $0\leq \alpha_{n}\leq 1/L$ and $0\leq \beta_{n}\leq 2/L-\alpha$,
 such that $\alpha_{n}+\beta_{n}=o(n\alpha_{n})$. We define $f_n= \pi^{2} /(4\alpha_{n} n^2)$ and we consider the sequence of quadratic functions \mbox{$f_n(\theta)=\frac{f_n\theta^2}{2}$}.
 We consider the iterate $(\eta_n)_n$ defined by our algorithm.
 We will show that
 \begin{equation*}
  \lim \alpha_{n}f_n(\eta_n)=\frac{\eta_{1}^{2}}{2}.
 \end{equation*}
  We have, from Lemma~\ref{lemma:stabcom},
  \begin{equation*}
  f_n(\eta_n)=\frac{\eta_n^2f_n}{2}=\frac{\eta_{1}^{2}\sin^{2}(\omega _{n}n)\rho_{n}^{2n}}{2\alpha_{n}(1-\frac{\pi^{2}(\alpha_{n}+\beta_{n})^{2}}{(4\alpha_{n} n)^{2}})}.
  \end{equation*}
  Moreover,
  \begin{equation*}
   \rho_{n}^{2n}=\left(1-\frac{\beta_{n} \pi^{2}}{4\alpha_{n} n^{2}}\right)^n=\exp\left(n\log\left(1-\frac{\beta_{n} \pi^{2}}{4\alpha_{n} n^{2}}\right)\right)= 1+o(1),
  \end{equation*}
  since $\frac{\beta_{n}}{\alpha_{n}n}=o(1)$.
   Also,  $1-\frac{\pi^{2}(\alpha_{n}+\beta_{n})^{2}}{(4\alpha_{n} n)^{2}}=1+o(1)$, since $\alpha_{n}+\beta_{n}=o(n\alpha_{n})$.
Moreover
  \begin{equation*}
  \sin(\omega_{n})=\frac{\sqrt{-\Delta_{n}}}{\rho_{n}}=\frac{\sqrt{f_{n}}\sqrt{\alpha_{n}-\frac{(\alpha_{n}+\beta_{n})^{2}}{4}f_{n}}}{\sqrt{1-\beta_{n} f_{n}}}={\pi}/(2n)+o(1/n),
  \end{equation*}
  thus $\omega_{n}={\pi}/(2n)+o(1/n)$ and $\sin(n\omega_{n})=1+o(1)$.

  \paragraph{Proof of the second lower-bound.}

 We consider now the situation where the second bound is active. Thus we take sequences  $(\alpha_n)$ and $(\beta_{n})$, such that  ${n\alpha_{n}}=o({\alpha_{n}+\beta_{n}})$. We define $g_n=\frac{2}{n(\alpha_n+\beta_n)}+\frac{4\alpha_n}{(\alpha_n+\beta_n)^2}$ and consider the sequence of quadratic functions $g_n(\theta)=\frac{g_n\theta^2}{2}$.
 We will show for the iterate $(\eta_n)$ defined by our algorithm that:
 \begin{equation*}
 \lim n(\alpha_n+\beta_n)(g_n(\theta_n)-g_n(\theta_*))=  \frac{\left(1-\exp(-2)\right)^2\Vert \theta_0-\theta_*\Vert^2}{4}.
 \end{equation*}
 We will use Lemma~\ref{lemma:stabreal}.
  We first have
     \begin{equation*}
    \Delta_{n}=\left(\frac{\alpha_n+\beta_n}{2}\right)^2g_n^2-\alpha_ng_n
    =g_n\left(\frac{\alpha_n+\beta_n}{2}\right)\frac{1}{n}.
   \end{equation*}
 Thus  $(n\Delta_{n})/g_n=\left(\frac{\alpha_n+\beta_n}{2}\right)$  and
 \begin{eqnarray}
  \sqrt{\Delta_n}&=&\sqrt{\left(\frac{1}{n}\right)^2+\frac{2\alpha_n}{n(\alpha_n+\beta_n)}}\nonumber\\
  &=&\frac{1}{n}\sqrt{1+\frac{2\alpha_n n}{\alpha_n+\beta_n}} \nonumber \\
  &=&\frac{1}{n}+\frac{\alpha_n}{\alpha_n+\beta_n}+o\left(\frac{\alpha_n}{\alpha_{n}+\beta_{n}}\right).\nonumber
 \end{eqnarray}
Moreover
  \begin{equation*}
r_{n}= 1-\frac{\alpha_n+\beta_n}{2}g_n
=1-\frac{1}{n} -\frac{2\alpha_n}{\alpha_n+\beta_n}.
 \end{equation*}
Thus \begin{equation*}
     r_+=1-\frac{\alpha_n}{\alpha_{n+}\beta_{n}}+o\left(\frac{\alpha_n}{\alpha_{n}+\beta_{n}}\right),
    \end{equation*}
    and \begin{equation*}
     r_+^n=\exp(n\log(r_+))=\exp\left(-\frac{n\alpha_n}{\alpha_{n}+\beta_{n}}\right)+o\left(\frac{n\alpha_n}{\alpha_{n}+\beta_{n}}\right)=1+o(1).
    \end{equation*}
 Furthermore
    \begin{equation*}
     r_-=1-\frac{2}{n}-\frac{3\alpha_n}{\alpha_{n+}\beta_{n}}+o\left(\frac{\alpha_n}{\alpha_{n}+\beta_{n}}\right),
    \end{equation*}
  and \begin{equation*}
     r_-^n=\exp(n\log(r_+))=\exp\left(-2-\frac{3\alpha_n n}{\alpha_{n}+\beta_{n}}\right)+o\left(\frac{n\alpha_n}{\alpha_{n}+\beta_{n}}\right)=\exp(-2)+o(1).
    \end{equation*}
    Thus
\begin{equation*}
 (r_+^n-r_-^n)^2=\left(1-\exp(-2)\right)^2+o(1).
\end{equation*}
Finally, we have:
    \begin{eqnarray}
      (\alpha_n+\beta_{n})n[g_n(\theta_n)-g_n(\theta_*)]&=&\frac{ \alpha_n+\beta_{n}}{2n}\Vert \theta_0-\theta_*\Vert^2\frac{[r_+^n-r_-^n]^2}{4\Delta_{n}/g_n}\nonumber\\
      &=&\frac{\Vert \theta_0-\theta_*\Vert^2}{4}{[r_+^n-r_-^n]^2}\nonumber \\
      &=&\frac{\Vert \theta_0-\theta_*\Vert^2}{4}{\left(1-\exp(-2)\right)^2}+o(1).\nonumber
    \end{eqnarray}

\section{Proofs of \mysec{noise}}\label{app:noise}

  \subsection{Proofs of Theorem \ref{cor:gennoise} and Theorem \ref{theo:gen}}
We decompose again vectors in an eigenvector basis of $H$ with $\eta_{n}^{i}=p_{i}^{\top}\eta_{n}$ and $\eps_{n}^{i}=p_{i}^{\top}\eps_{n}$:  
\begin{equation*}
 \eta_{n+1}^{i}=(1-\al h_{i})\eta_n^{i} +(1-\be h_{i})(\eta_n^{i}-\eta_{n-1}^{i})+(n\alpha+\beta) \eps_{n+1}^{i}.
\end{equation*}
We denote by $\xi_{n+1}^{i}=\begin{pmatrix}
              [n\al+\be] \eps_{n+1}^{i}\\
              0
             \end{pmatrix}$
and we have the reduced equation:
\begin{equation*}
 \Theta_{n+1}^{i}=F_{i}\Theta^i_n+\xi_{n+1}^{i}.
\end{equation*}
Unfortunately $F_{i}$ is not Hermitian and this formulation will not be convenient for calculus. 
Without loss of generality, we assume \mbox{$r_{i}^{-} \neq r_{i}^{+}$} even if it means having \mbox{$r_{i}^{-}-r_{i}^{+}$}  goes to $0$ in the final bound.
Let $Q_{i}=\begin{pmatrix}
             r_{i}^{-} &r_{i}^{+} \\
              1&1
             \end{pmatrix}$ be the transfer matrix of $F_{i}$, i.e., $F_{i}=Q_{i}D_{i}Q_{i}^{-1}$ with \mbox{$D_i=\begin{pmatrix}
             r_{i}^{-} &0 \\
              0&r_{i}^{+}
             \end{pmatrix}$}
             and 
             \mbox{$Q_{i}^{-1}=\frac{1}{r_{i}^{-}-r_{i}^{+}}\begin{pmatrix}
             1 &-r_{i}^{+} \\
              -1&r_{i}^{-}
             \end{pmatrix}$}.
 We can reparametrize the problem in the following way: 
             \begin{eqnarray}
 Q_{i}^{-1}\Theta_{n+1}^{i}&=&Q_{i}^{-1}F_{i}\Theta_n ^{i}+Q_{i}^{-1} \xi_{n+1}^{i} \nonumber \\
 &=&Q_{i}^{-1}F_{i}Q_{i} Q_{i}^{-1}\Theta^{i}_n+Q_{i}^{-1} \xi_{n+1}^{i} \nonumber \\
 &=& D_{i} (Q_{i}^{-1}\Theta_n^{i})+Q_{i}^{-1} \xi_{n+1}^{i}. \nonumber
\end{eqnarray}  
With $ \tilde \Theta_{n}^{i}=Q_{i}^{-1}\Theta_{n}^{i}$ and $\tilde \xi_n^{i}=Q_{i}^{-1}\xi _n^{i}$ we now have:
\begin{equation}\label{eq:xi}
 \tilde \Theta_{n+1}^{i}=D_{i}\tilde \Theta^{i}_n+\tilde \xi^{i}_{n+1},
\end{equation}         
with now $D_{i}$ Hermitian (even diagonal).

Thus it is easier to tackle using standard techniques for stochastic approximation  \citep[see, e.g.,][]{pj,bm2}:
\begin{equation*}
 \tilde \Theta^{i}_n=D_{i}^n\tilde\Theta_0^{i}+\sum_{k=1}^nD_{i}^{n-k}\tilde \xi_k^{i}.
\end{equation*}           
 Let $M_{i}=\begin{pmatrix} h_{i}^{1/2}& h_{i}^{1/2}\\0&0\end{pmatrix}$, we then get using standard martingale square moment inequalities, 
 since for $n\neq m$, $\eps_{n}^i$ and $\eps_{m}^i$ are uncorrelated (i.e., \mbox{$\EE[\eps_{n}^i\eps_{m}^i]=0$}):
\begin{equation*}
 \EE \Vert M_{i} \tilde \Theta_n^{i} \Vert ^2 = \Vert M_{i}D_{i}^n\tilde\Theta_0^{i} \Vert ^2+\EE\sum_{k=1}^n \Vert M_{i}D_{i}^{n-k}\tilde \xi_k^{i} \Vert^2.
\end{equation*}
This is a bias-variance decomposition; the left term only depends on the initial condition and the right term only depends  on the noise process.

We have with $M_{i}=\begin{pmatrix} h_{i}^{1/2}& h_{i}^{1/2}\\0&0\end{pmatrix}$, $M_{i}Q_i^{-1}=\begin{pmatrix} 0&h_{i}^{1/2}\\0&0\end{pmatrix}$, and \mbox{$M_{i}\tilde \Theta_{n}^{i}=\begin{pmatrix} \sqrt{h_{i}} \eta_{n}^{i}\\0\end{pmatrix}$}.
Thus, we have access to the function values through:
    \begin{equation*}
\Vert M_{i}\tilde \Theta_{n}^{i}\Vert^{2}=h_{i}(\eta_{n}^i)^{2}.
\end{equation*} 
Moreover we have  $\Theta_{0}^{i}=\begin{pmatrix}
             \phi^i_{1}/(r_{i}^{-}-r_{i}^{+})\\
              -\phi^i_{1}/(r_{i}^{-}-r_{i}^{+})
             \end{pmatrix}$.
Thus
    \begin{equation*}
\Vert M_{i}D_{i}^{n}\tilde \Theta_{0}^{i}\Vert^{2}=(\phi_{1}^{i})^{2}h_{i}\frac{\left((r_{i}^{+})^{n}-(r_{i}^{-})^{n}\right)^{2}}{(r_{i}^{+}-r_{i}^{-})^{2}}.
\end{equation*}         
This is the bias term we have studied in \mysec{generalbound} which we bound with Theorem \ref{theo:bounddet}. The variance term is controlled by the next proposition.

\begin{proposition}\label{prop:noise}
With $\EE[(\eps_n^i)^2]=c_i$ for all $n\in \NN$,  for $\alpha\leq{1}/{h_{i}}$ and $0\leq \beta\leq {2}/{h_i}-\alpha$ , we have
  \begin{multline*}
  \frac{1}{n^2}\EE\sum_{k=1}^n \Vert M_{i}D_{i}^{n-k}\tilde \xi_k^{i} \Vert^2\leq 
 \\  \min \left \{ \frac{2(\al n+\be)^2}{\alpha \beta(4-(\alpha+2\beta)h_i) n^{2}}\frac{c_{i}}{h_i},\frac{16(n\alpha+\beta)^2}{n({\alpha+\beta})^2}\frac{c_{i}}{h_i},  2\frac{(\alpha n+\beta)^{2}}{n\alpha} c_{i},\frac{8(n\alpha+\beta)^2}{{\alpha+\beta}}c_{i}\right\}.
\end{multline*}
  \end{proposition}
The last two bounds prove Theorem  \ref{cor:gennoise}.

We note that if we restrict $\beta$ to $\beta\leq 3/(2h_i)-\alpha/2$, then $4-(\alpha+2\beta)h_i \geq 1$ and the first bound of Proposition \ref{prop:noise} is simplified to 
\mbox{$\frac{2(\al n+\be)^2}{\alpha \beta n^{2}}\frac{c_{i}}{h_i}$}.
This allows to conclude to prove Theorem \ref{theo:gen}.

\subsection{Proof of Corollary \ref{cor:ob}}
We let $\nu=\frac{\Vert \theta_0-\theta_*\Vert}{\sqrt{L\tr(CH^{-1})}}$ and consider three different regimes depending on $\nu$ and $L$.

If $\nu<1/L$, we have $\nu/N<1/L$ and thus $\alpha=\nu/N$ and $\beta=\nu$. Therefore
\BEA
\frac{\Vert \theta_0-\theta_*\Vert^2}{N^2\alpha}+\frac{(\alpha N+\beta)^2}{\alpha\beta N^2}\tr ( C H^{-1})&=&  \frac{\Vert \theta_0-\theta_*\Vert^2}{\nu N}+   \frac{4\tr ( C H^{-1})}{N} \nonumber \\
&\leq&\frac{\sqrt{L\tr(CH^{-1})}\Vert \theta_0-\theta_*\Vert}{N}+   \frac{4\tr ( C H^{-1})}{N} \nonumber \\
&\leq&\frac{5\tr ( C H^{-1})}{N}, \nonumber
\EEA
where we have used $\sqrt{L}\Vert \theta_0-\theta_*\Vert<\sqrt{\tr(CH^{-1})}$ since $\nu<1/L$.

If $\nu>1/L$ and $\nu<N /L$, we have $\alpha=\nu/N$ and $\beta=1/L$. Therefore
\BEA
\frac{\Vert \theta_0-\theta_*\Vert^2}{N^2\alpha}+\frac{(\alpha N+\beta)^2}{\alpha\beta N^2}\tr ( C H^{-1})&\leq&  \frac{\Vert \theta_0-\theta_*\Vert^2}{\nu N}+   \frac{4\tr ( C H^{-1})}{L\nu N} \nonumber \\
&\leq&\frac{\sqrt{L\tr(CH^{-1})}\Vert \theta_0-\theta_*\Vert}{N}+   \frac{4\tr ( C H^{-1})}{N} \nonumber \\
&\leq&\frac{5\sqrt{L\tr(CH^{-1})}\Vert \theta_0-\theta_*\Vert}{N}, \nonumber
\EEA
where we have used $\sqrt{L}\Vert \theta_0-\theta_*\Vert>\sqrt{\tr(CH^{-1})}$ since $\nu>1/L$.

If $\nu>N/L$, we have $\alpha=1/L$ and $\beta=1/L$. Therefore
\BEA
\frac{\Vert \theta_0-\theta_*\Vert^2}{N^2\alpha}+\frac{(\alpha (N-1)+\beta)^2}{\alpha\beta N^2}\tr ( C H^{-1})&=&  \frac{L\Vert \theta_0-\theta_*\Vert^2}{N^2}+   \tr ( C H^{-1}) \nonumber \\
&\leq& \frac{L\Vert \theta_0-\theta_*\Vert^2}{N^2}+    \frac{L\Vert \theta_0-\theta_*\Vert^2}{N^2} \nonumber \\
&\leq& \frac{2L\Vert \theta_0-\theta_*\Vert^2}{N^2}, \nonumber
\EEA
where we have used that the real bound in Proposition \ref{prop:noise} is in fact in $(N-1)\alpha+\beta$, (see Lemma \ref{lemma:sumnoise}) and that $\tr ( C H^{-1})<\frac{L\Vert \theta_0-\theta_*\Vert^2}{N^2}$
since $\nu>N /L$.
       
\subsection{Proof of Proposition \ref{prop:noise}}

 \subsubsection{Proof outline}
 
 To prove Proposition \ref{prop:noise} we will use Lemmas \ref{lemma:sumnoise}, \ref{lemma:ineq} and \ref{lemma:loose}, that are stated and proved in \mysec{techlemma}.
 
 We want to bound $\EE[\sum_{k=1}^n \Vert M_{i}D_{i}^{n-k}\tilde \xi_k^{i} \Vert^2]$ and according to Lemma \ref{lemma:sumnoise}, we have an explicit expansion using the roots of the characteristic polynomial:
\begin{equation*}
\EE \Vert M_{i}D_{i}^{k}\tilde \xi^{i}_k \Vert^2=h_{i} {((k-1)\al+\be)^2} \EE[{(\eps^{i})}^2] \frac{[(r_{i}^{-})^{n-k}-(r_{i}^{+})^{n-k}]^2}{(r_{i}^{-}-r_{i}^{+})^{2}}.
\end{equation*}
Thus, by bouding $(k-1)\al+\be$ by $(n-1)\al+\be$, we get
\begin{equation}\label{eq:sumroo}
\EE\sum_{k=1}^n \Vert M_{i}D_{i}^{n-k}\tilde \xi_k^{i} \Vert^2\leq h_{i} ((n-1)\al+\be)^2{\EE[{\eps^{i}}^2] } \sum_{k=1}^n \frac{[(r_{i}^{-})^{n-k}-(r_{i}^{+})^{n-k}]^2}{(r_{i}^{-}-r_{i}^{+})^{2}}.
\end{equation}
Then, we have from Lemma \ref{lemma:ineq} the inequality:
 \begin{equation*}
   \sum_{k=1}^n \frac{[(r_{i}^{-})^{k}-(r_{i}^{+})^{k}]^2}{[(r_{i}^{-})-(r_{i}^{+})]^2}\leq\frac{2-\beta h_i}{4\alpha \beta h_i^2(1-(\frac{1}{4}\alpha+\frac{1}{2}\beta)h_i)}.
\end{equation*}
Therefore
\begin{equation*}
\EE\sum_{k=1}^n \Vert M_{i}^{1/2}D_{i}^{n-k}\tilde \xi_k^{i} \Vert^2\leq \frac{\EE[{\eps^{i}}^2]}{h_i} \frac{((n-1)\al+\be)^2}{4\alpha \beta}\frac{2-\beta h_i}{(1-(\frac{1}{4}\alpha+\frac{1}{2}\beta)h_i)}.
\end{equation*}
This allows to prove the first part of the bound.  The other parts are much simpler and are done in Lemma \ref{lemma:loose}.
Thus, adding these bounds gives  for $\alpha\leq1/h_{i}$ and $0\leq \beta\leq 2/h_i-\alpha$:
  \begin{multline*}
  \frac{1}{n^{2}}\EE\sum_{k=1}^n \Vert M_{i}D_{i}^{n-k}\tilde \xi_k^{i} \Vert^2\leq 
  \\\min \left \{ \frac{2(\al{(n-1)}+{\be})^2}{\alpha \beta n^{2}(4-(\alpha+2\beta)h_i)}\frac{c}{h_i},\frac{16((n-1)\alpha+\beta)^2}{n({\alpha+\beta})^2}\frac{c}{h_i}, 2\frac{(\alpha (n-1)+\beta)^{2}}{n\alpha} c_{i},\frac{8((n-1)\alpha+\beta)^2}{{\alpha+\beta}}c_{i}\right\}.
  \end{multline*}

\subsubsection{Some technical Lemmas}\label{sec:techlemma}
 
We first compute an explicit expansion of the noise term as a function of the eigenvalues of the dynamical system.  
\begin{lemma}\label{lemma:sumnoise}
For all $\alpha\leq 1/h_i$ and $0\leq \beta\leq 2/h_i-\alpha$ we have
\begin{equation*}
\EE \Vert M_{i}D_{i}^{k}\tilde \xi^{i}_k \Vert^2=h_{i} {((k-1)\al+\be)^2} \EE[{(\eps^{i})}^2] \frac{[(r_{i}^{-})^{n-k}-(r_{i}^{+})^{n-k}]^2}{(r_{i}^{-}-r_{i}^{+})^{2}}.
\end{equation*}
\end{lemma}
\begin{proof}
We first turn the Euclidean norm into a trace, using that $\tr[AB]=\tr[BA]$ for two matrices $A$ and $B$ and that $\tr[x]=x$ for a real $x$.
 \begin{equation}\label{eq:mdxi}
 \EE \Vert M_{i}D_{i}^{n-k}\tilde \xi^{i}_k \Vert^2=\text{Tr} {D_{i}^{n-k}}{M_{i}}^\top M_{i}D_{i}^{n-k}\EE [\tilde \xi^{i}_k {(\tilde {\xi}^{i}_k)}^\top],
\end{equation}
This enables us to separate the noise term from the rest of the formula. Then we compute the latter from the definition of $\tilde \xi^{i}_k$ in \eq{xi} :
\begin{equation*}
\EE [\tilde \xi^{i}_k {(\tilde {\xi}^{i}_k)}^\top]
=\frac{((k-1)\al+\be)^2}{(r_{i}^{-}-r_{i}^{+})^{2}}\EE[{(\eps^{i})}^2]
\begin{pmatrix}
  1&-1\\-1&1 \end{pmatrix}.
\end{equation*}
And the first part of \eq{mdxi} is equal to:
\begin{equation*}
 {D_{i}^{n-k}} {M_{i}}^\top M_{i}D_{i}^{n-k}=h_{i}
 \begin{pmatrix}
  {(r_{i}^{-})}^{2(n-k)}&{(r_{i}^{-})^{(n-k)}-(r_{i}^{+})}^{(n-k)}\\
{(r_{i}^{-})^{(n-k)}-(r_{i}^{+})}^{(n-k)}&{(r_{i}^{+})}^{2(n-k)} \end{pmatrix},
\end{equation*}
because \mbox{$D_i=\begin{pmatrix}
             r_{i}^{-} &0 \\
              0&r_{i}^{+}
             \end{pmatrix}$} and \mbox{$M_{i}=\begin{pmatrix} h_{i}^{1/2}& h_{i}^{1/2}\\0&0\end{pmatrix}$}.
Therefore:
\begin{equation*}
\EE \Vert M_{i}D_{i}^{n-k}\tilde \xi^{i}_k \Vert^2=h_{i} \frac{((k-1)\al+\be)^2}{(r_{i}^{-}-r_{i}^{+})^{2}} \EE[{\eps^{i}}^2] [(r_{i}^{-})^{n-k}-(r_{i}^{+})^{n-k}]^2.
\end{equation*}
\end{proof}      

In the following leamma, we bound a certain  sum of  powers of the roots.        
\begin{lemma}\label{lemma:ineq}
      For all $\alpha\leq 1/h_i$ and $0\leq \beta\leq 2/h_i-\alpha$ we have
 \begin{equation*}
   \sum_{k=1}^n \frac{\big[(r_{i}^{-})^{k}-(r_{i}^{+})^{k}]^2}{[(r_{i}^{-})-(r_{i}^{+})\big]^2}\leq\frac{2-\beta h_i}{4\alpha \beta h_i^2(1-(\frac{1}{4}\alpha+\frac{1}{2}\beta)h_i)}.
\end{equation*}
\end{lemma}
We first note that when the two roots become close, the denominator and the numerator will go to zero, which prevents from bounding the numerator easily.
We also note that this bound is very tight since the difference between the two terms goes to zero when $n$ goes to infinity.
\begin{proof}
We first expand the square of the difference of the powers of the roots and compute their sums.
 \begin{eqnarray}
   \sum_{k=1}^n {\big[(r_{i}^{-})^{k}-(r_{i}^{+})^{k}\big]^2}&=& \sum_{k=1}^n\big[{r_{i}^{+}}^{2k}+{r_{i}^{-}}^{2k}-2(r_{i}^{+}r_{i}^{-})^{k}\big] \nonumber \\
   &=&\frac{1-{r_{i}^{+}}^{2n}}{1-{r_{i}^{+}}^{2}}+\frac{1-{r_{i}^{-}}^{2n}}{1-{r_{i}^{-}}^{2}}-2\frac{1-(r_{i}^{+}r_{i}^{-})^{n}}{1-(r_{i}^{+}r_{i}^{-})} \nonumber \\
   &=&\frac{1}{1-{r_{i}^{+}}^{2}}+\frac{1}{1-{r_{i}^{-}}^{2}}-\frac{2}{1-(r_{i}^{+}r_{i}^{-})} -\bigg[\frac{{r_{i}^{+}}^{2n}}{1-{r_{i}^{+}}^{2}}+\frac{{r_{i}^{-}}^{2n}}{1-{r_{i}^{-}}^{2}}-2\frac{(r_{i}^{+}r_{i}^{-})^{n}}{1-(r_{i}^{+}r_{i}^{-})}\bigg] \nonumber \\
     &=&\frac{1}{1-{r_{i}^{+}}^{2}}+\frac{1}{1-{r_{i}^{-}}^{2}}-\frac{2}{1-(r_{i}^{+}r_{i}^{-})} -I_n,\nonumber
  \end{eqnarray}
  with $I_n=\bigg[\frac{{r_{i}^{+}}^{2n}}{1-{r_{i}^{+}}^{2}}+\frac{{r_{i}^{-}}^{2n}}{1-{r_{i}^{-}}^{2}}-2\frac{(r_{i}^{+}r_{i}^{-})^{n}}{1-(r_{i}^{+}r_{i}^{-})}\bigg] $.
  
This sum is therefore equal to the sum of one term we will compute explicitly and one other term which will go to zero. 
We have for the first term:
 \begin{multline*}
  \frac{1}{1-{r_{i}^{+}}^{2}}+\frac{1}{1-{r_{i}^{-}}^{2}}-\frac{2}{1-(r_{i}^{+}r_{i}^{-})}=\frac{(1-{r_{i}^{-}}^{2})(1-(r_{i}^{+}r_{i}^{-}))-(1-{r_{i}^{-}}^{2})(1-{r_{i}^{+}}^{2})}{(1-{r_{i}^{+}}^{2})(1-{r_{i}^{-}}^{2})(1-(r_{i}^{+}r_{i}^{-}))}
 \\+ \frac{(1-{r_{i}^{+}}^{2})(1-(r_{i}^{+}r_{i}^{-}))-(1-{r_{i}^{-}}^{2})(1-{r_{i}^{+}}^{2})}{(1-{r_{i}^{+}}^{2})(1-{r_{i}^{-}}^{2})(1-(r_{i}^{+}r_{i}^{-}))},
 \end{multline*}
with
  \begin{eqnarray}
(1-{r_{i}^{-}}^{2})(1-(r_{i}^{+}r_{i}^{-}))-(1-{r_{i}^{-}}^{2})(1-{r_{i}^{+}}^{2})&=&(1-{r_{i}^{-}}^{2})[(1-(r_{i}^{+}r_{i}^{-}))-(1-{r_{i}^{+}}^{2})]\nonumber \\
&=&r_{i}^{+}(1-{r_{i}^{-}}^{2})(r_{i}^{+}-r_{i}^{-}),\nonumber
    \end{eqnarray}
and 
\begin{equation*}
 (1-{r_{i}^{+}}^{2})(1-(r_{i}^{+}r_{i}^{-}))-(1-{r_{i}^{-}}^{2})(1-{r_{i}^{+}}^{2})=-r_{i}^{-}(1-{r_{i}^{-}}^{2})(r_{i}^{+}-r_{i}^{-}),
\end{equation*}
and 
  \begin{eqnarray}
 r_{i}^{+}(1-{r_{i}^{-}}^{2})(r_{i}^{+}-r_{i}^{-})-r_{i}^{-}(1-{r_{i}^{-}}^{2})(r_{i}^{+}-r_{i}^{-})&=&(r_{i}^{+}-r_{i}^{-})[r_{i}^{+}(1-{r_{i}^{-}}^{2})-r_{i}^{-}(1-{r_{i}^{-}}^{2})]\nonumber \\
 &=&(r_{i}^{+}-r_{i}^{-})[r_{i}^{+}-r_{i}^{-}+r_{i}^{+}r_{i}^{-}(r_{i}^{+}-r_{i}^{-})]\nonumber \\
 &=&(r_{i}^{+}-r_{i}^{-})^2[1+r_{i}^{+}r_{i}^{-}]. \nonumber
    \end{eqnarray}
Therefore the first term is equal to:
\begin{equation*}
  \frac{1}{1-{r_{i}^{+}}^{2}}+\frac{1}{1-{r_{i}^{-}}^{2}}-\frac{2}{1-(r_{i}^{+}r_{i}^{-})}=\frac{(r_{i}^{+}-r_{i}^{-})^2[1+r_{i}^{+}r_{i}^{-}]}{(1-{r_{i}^{+}}^{2})(1-{r_{i}^{-}}^{2})(1-(r_{i}^{+}r_{i}^{-}))},
\end{equation*}
and the sum can be expanded as:
\begin{equation*}
  \sum_{k=1}^n \frac{[(r_{i}^{-})^{k}-(r_{i}^{+})^{k}]^2}{[r_{i}^{-}-r_{i}^{+}]^2}=\frac{[1+r_{i}^{+}r_{i}^{-}]}{(1-{r_{i}^{+}}^{2})(1-{r_{i}^{-}}^{2})(1-(r_{i}^{+}r_{i}^{-}))}-J_n,
\end{equation*}
with $J_n=\frac {I_n}{[(r_{i}^{-})-(r_{i}^{+})]^2}$.

Then we simplify the first term of this sum using the explicit values of the roots.
We recall $r_{i}^{\pm}=r_i\pm \sqrt{\Delta_i}=1-\frac{\alpha+\beta}{2}h_i\pm\sqrt{\left(\frac{\alpha+\beta}{2}\right)^2h_i^2-\alpha h_i}$, therefore
  \begin{eqnarray}
   r_{i}^{+}r_{i}^{-}&=&r_i^2-\Delta_i^2\nonumber \\
   &=&\left(1-\left(\frac{\alpha+\beta}{2}\right)h_i\right)^2-\left[\left(\frac{\alpha+\beta}{2}\right)h_i\right]^2+\alpha h_i \nonumber \\
   &=&1-\beta h_i,\nonumber 
  \end{eqnarray}
  and 
    \begin{eqnarray}
     (1-{r_{i}^{+}}^{2})(1-{r_{i}^{-}}^{2})&=&[(1-{r_{i}^{-}})(1-{r_{i}^{+}})][(1+{r_{i}^{+}})(1+{r_{i}^{+}})]\nonumber \\
     &=&[(1-r_i+\sqrt{\Delta_i})((1-r_i-\sqrt{\Delta_i})] [(1+r_i+\sqrt{\Delta_i})((1+r_i-\sqrt{\Delta_i})]\nonumber \\
     &=&[(1-r_i)^2-\Delta_i][(1+r_i)^2-\Delta_i][(1-r_i)^2-\Delta_i] \nonumber \\
     &=&4\alpha h_i\left(1-\left(\frac{1}{4}\alpha+\frac{1}{2}\beta\right)h_i\right).\nonumber
    \end{eqnarray}
 Thus 
 \begin{equation*}
  \sum_{k=1}^n \frac{[(r_{i}^{-})^{k}-(r_{i}^{+})^{k}]^2}{[(r_{i}^{-})-(r_{i}^{+})]^2}=\frac{2-\beta h_i}{4\alpha \beta h_i^2(1-(\frac{1}{4}\alpha+\frac{1}{2}\beta)h_i)}-J_n.
 \end{equation*}
 
 Even if $J_n$ will be asymptotically small, we want a non-asymptotic bound, thus we will show that $J_n$ is always positive.

 In the real case $[(r_{i}^{-})-(r_{i}^{+})]^2\geq 0$ and using $a^2+b^2\geq 2ab$, for all $(a,b)\in\RR^2$, we have 
\begin{equation*}
 \frac{{r_{i}^{+}}^{2n}}{1-{r_{i}^{+}}^{2}}+\frac{{r_{i}^{-}}^{2n}}{1-{r_{i}^{-}}^{2}}\geq 2\frac{({r_{i}^{+}}{r_{i}^{-}})^n}{\sqrt{(1-{r_{i}^{+}}^{2})(1-{r_{i}^{-}}^{2})}},
\end{equation*}
and using ${r_{i}^{+}}^{2}+{r_{i}^{-}}^{2}\geq 2{r_{i}^{-}}{r_{i}^{-}}$
we have 
\begin{equation*}
 \sqrt{(1-{r_{i}^{+}}^{2})(1-{r_{i}^{-}}^{2})}\leq 1-(r_{i}^{+}r_{i}^{-}),
\end{equation*}
since 
  \begin{eqnarray}
(1-{r_{i}^{+}}^{2})(1-{r_{i}^{-}}^{2})-[1-(r_{i}^{+}r_{i}^{-})]^2&=&1-{r_{i}^{+}}^{2}-{r_{i}^{-}}^{2}+(r_{i}^{+}r_{i}^{-})^2-1+2r_{i}^{+}r_{i}^{-}-(r_{i}^{+}r_{i}^{-})^2\nonumber \\
&=& 2r_{i}^{+}r_{i}^{-}-{r_{i}^{+}}^{2}-{r_{i}^{-}}^{2}\nonumber \\
&\leq&0.\nonumber
    \end{eqnarray}
    Thus 
     \begin{equation*}
\frac{{r_{i}^{+}}^{2n}}{1-{r_{i}^{+}}^{2}}+\frac{{r_{i}^{-}}^{2n}}{1-{r_{i}^{-}}^{2}}-2\frac{(r_{i}^{+}r_{i}^{-})^{n}}{1-(r_{i}^{+}r_{i}^{-})}  \geq 0.
 \end{equation*}
 and $J_n\geq0$ in the real case.

 In the complex case, $[(r_{i}^{-})-(r_{i}^{+})]^2\leq 0$, and using $z^2+\bar z^2\leq 2z\bar z$ for all $z\in \CC$, we have 
 \begin{equation*}
 \frac{{r_{i}^{+}}^{2n}}{1-{r_{i}^{+}}^{2}}+\frac{{r_{i}^{-}}^{2n}}{1-{r_{i}^{-}}^{2}}\leq 2\frac{({r_{i}^{+}}{r_{i}^{-}})^n}{\sqrt{(1-{r_{i}^{+}}^{2})(1-{r_{i}^{-}}^{2})}},
\end{equation*}
 and using ${r_{i}^{+}}^{2}+{r_{i}^{-}}^{2}\leq 2{r_{i}^{-}}{r_{i}^{-}}$
we have 
\begin{equation*}
 \sqrt{(1-{r_{i}^{+}}^{2})(1-{r_{i}^{-}}^{2})}\geq 1-(r_{i}^{+}r_{i}^{-}).
\end{equation*}   
Thus 
     \begin{equation*}
\frac{{r_{i}^{+}}^{2n}}{1-{r_{i}^{+}}^{2}}+\frac{{r_{i}^{-}}^{2n}}{1-{r_{i}^{-}}^{2}}-2\frac{(r_{i}^{+}r_{i}^{-})^{n}}{1-(r_{i}^{+}r_{i}^{-})}  \leq 0.
 \end{equation*}
and $J_n\geq0$ in the complexe case.

Therefore we always have:
\begin{equation*}
 J_n\geq 0, 
\end{equation*}
and 
\begin{equation*}
   \sum_{k=1}^n \frac{[(r_{i}^{-})^{k}-(r_{i}^{+})^{k}]^2}{[(r_{i}^{-})-(r_{i}^{+})]^2}\leq\frac{2-\beta h_i}{4\alpha \beta h_i^2(1-(\frac{1}{4}\alpha+\frac{1}{2}\beta)h_i)}.
\end{equation*}
\end{proof}

However we can also bound roughly \eq{sumroo} using Theorem \ref{theo:bounddet} since we recall we have $\eta_n^i=\frac{[(r_{i}^{-})^{n-k}-(r_{i}^{+})^{n-k}]^2}{(r_{i}^{-}-r_{i}^{+})^{2}}$.
This gives us the following lemma which enables to prove the second part of Proposition \ref{prop:noise}.
\begin{lemma}\label{lemma:loose}
      For all $\alpha\leq 1/h_i$ and $0\leq \beta\leq 2/h_i-\alpha$ we have
  \begin{equation*}
   \EE\sum_{k=1}^n \Vert M_{i}^{1/2}D_{i}^{n-k}\tilde \xi_k^{i} \Vert^2\leq\EE[{(\eps^{i})}^2]n((n-1)\alpha+\beta)^2\min\left\{\frac{2}{\alpha},\frac{8n}{\alpha+\beta},\frac{16}{h_i(\alpha+\beta)^2}\right\}.
  \end{equation*}
 \end{lemma}
 
 \begin{proof}
 From Lemma \ref{lemma:sumnoise}, we get
  \begin{eqnarray}
   \EE\sum_{k=1}^n \Vert M_{i}^{1/2}D_{i}^{n-k}\tilde \xi_k^{i} \Vert^2&=&h_{i} {\EE[{(\eps^{i})}^2] } \sum_{k=1}^n ((k-1)\al+\be)^2\frac{[(r_{i}^{-})^{n-k}-(r_{i}^{+})^{n-k}]^2}{(r_{i}^{-}-r_{i}^{+})^{2}} \nonumber \\
   &\leq&h_i\EE[{(\eps^{i})}^2]((n-1)\alpha+\beta)^2n\min\left\{\frac{2}{\alpha h_i},\frac{8n}{({\alpha+\beta})h_i},\frac{16}{({\alpha+\beta})^2h_i^2}\right\}\nonumber\\
   &\leq&\EE[{(\eps^{i})}^2]n((n-1)\alpha+\beta)^2\min\left\{\frac{2}{\alpha},\frac{8n}{\alpha+\beta},\frac{16}{h_i(\alpha+\beta)^2}\right\}. \nonumber
  \end{eqnarray}
\end{proof}

\section{Comparison with additional other algorithms}\label{app:comparison}

\subsection{Summary}

 When the objective function $f$ is quadratic and for correct choices of step-sizes, the AC-SA algorithm of \cite{lan}, the SAGE algorithm of \cite{sage} and the Accelerated RDA algorithm of \cite{xiao} are all equivalent to:
   \begin{equation*}
    \tnp=[I-\delta_{n+1}H_{n+1}]\theta_n+\frac{n-2}{n+1}[I-\delta_{n+1}H_{n+1}](\tn-\tnm)+\delta_{n+1}\eps_{n+1},
   \end{equation*}
where we use $H_{n}\theta+\eps_n$ as an unbiased estimate of the gradient and $\delta_{n}$ as step-size which values will be specified later.

\cite{lan} and \cite{sage} only consider bounded cases by projecting their iterates on a bounded space. \cite{xiao} deals with the unbounded case and prove the following convergence result:
\begin{theorem} \citep[][Theorem 6]{xiao}.
With $\EE[\eps_{n}\otimes \eps_{n}]=C$, for step-size $\delta_n\leq \frac{n-1}{n}\gamma$ with $\gamma\leq1/L$, we have 
 \begin{equation*}
  \EE f(\theta_n)-f(\theta_*)\leq \frac{4\Vert\theta_0-\theta_*\Vert^2}{n^2\gamma}+\frac{n\gamma \sigma^2\tr C}{3}.
 \end{equation*}
\end{theorem}
This result is significantly more general than ours since it is valid for composite optimization and  general noise on the gradients. 

We now present the different algorithms and show they all share the same form.

  \subsection{AC-SA}
  
  \begin{lemma}
    AC-SA algorithm with step size $\gamma_n$ and $\beta_n$ and gradient estimate $H_{n+1}\theta_n+\eps_{n+1}$ is equivalent to:
  \begin{equation*}
   \tnp=(I-\frac{\gamma_n}{\beta_n}H_{n+1})\tn+\frac{\beta_{n-1}-1}{\beta_n}(I-\frac{\gamma_n}{\beta_n}H_{n+1})(\tn-\tnm)+\frac{\gamma_n}{\beta_n} \eps_{n+1}.
  \end{equation*}
  \end{lemma}

\begin{proof}
 We recall the general \textbf{AC-SA algorithm}:
 \begin{itemize}
  \item Let the initial points $x_1^{ag}=x_1$, and the step-sizes $\{\beta_n\}_{n\leq1}$ and $\{\gamma_n\}_{n\leq1}$ be given. 
  
  Set $n=1$
  \item \textbf{Step 1}.
  Set $x_n^{md}=\beta_n^{-1}x_n+(1-\beta_n^{-1})x_n^{ag}$,
  \item \textbf{Step 2}.
  Call the Oracle for computing $G(x_n^{md},\xi_n)$ where $\EE[G(x_n^{md},\xi_n)]=f'(x_n^{md})$.
  
  Set 
  \begin{equation*}
   x_{n+1}=x_n-\gamma_n G(x_n^{md},\xi_n),
  \end{equation*}
  \begin{equation*}
   x_{n+1}^{ag}=\beta_n^{-1}x_{n+1}+(1-\beta_n^{-1})x_n^{ag},
  \end{equation*}
  \item \textbf{Step 3}.
  Set $n\rightarrow n+1$ and go to step 1.
 \end{itemize}

  When $f$ is quadratic we will have $G(x_n^{md},\xi_n)=H_{n+1} x_n^{md}-\eps_{n+1}$,
  thus $x_{n+1}=x_n-\gamma_n H_{n+1} x_n^{md}+\gamma_n \eps_{n+1}$, and:
    \begin{eqnarray}
x_{n+1}^{ag}&=&\beta_n^{-1}x_{n+1}+(1-\beta_n^{-1})x_n^{ag}\nonumber \\
&=&\beta_n^{-1}(x_n-\gamma_n H_{n+1} x_n^{md}+\gamma_n \eps_{n+1})+(1-\beta_n^{-1})x_n^{ag}\nonumber \\
&=&\beta_n^{-1}(\beta_nx_n^{md}+(1-\beta_n)x_n^{ag}-\gamma_n H_{n+1} x_n^{md}+\gamma_n \eps_{n+1})+(1-\beta_n^{-1})x_n^{ag}\nonumber \\
&=&x_n^{md}-\frac{\gamma_n}{\beta_n} H_{n+1} x_n^{md}+\frac{\gamma_n}{\beta_n} \eps_{n+1},\nonumber 
  \end{eqnarray}
  and 
      \begin{eqnarray}
       x_n^{md}&=&\beta_n^{-1}x_n+(1-\beta_n^{-1})x_n^{ag}\nonumber \\
       &=&\beta_n^{-1}\beta_{n-1}x_n^{ag}+\beta_n^{-1}(1-\beta_{n-1})x_{n-1}^{ag}+(1-\beta_n^{-1})x_n^{ag}\nonumber \\
       &=&x_n^{ag}+\frac{\beta_{n-1}-1}{\beta_n}[x_{n}^{ag}-x_{n-1}^{ag}].\nonumber
      \end{eqnarray}
 These give the result for $\theta_n=x_n^{ag}$.
\end{proof}

\subsection{SAGE}

\begin{lemma}
 The algorithm SAGE with step-sizes $L_n$ and $\alpha_n$ is equivalent to: 
 \begin{equation*}
  \theta_{n+1}=(I-1/L_{n+1} H_{n+1})\theta_n+(1-\alpha_{n})\frac{\alpha_{n+1}}{\alpha_{n}}[I-1/L_{n+1} H_{n+1}](\tn-\tnm)+1/L_{n+1}\eps_{n+1}.
 \end{equation*}
\end{lemma}

\begin{proof}
 We recall the general \textbf{SAGE algorithm}:
 \begin{itemize}
  \item Let the initial points $x_0=z_0=0$, and the step-sizes $\{\beta_n\}_{n\leq1}$ and $\{L_{n}\}_{n\leq1}$ be given. 
  
  Set $n=1$
  \item \textbf{Step 1}.
  Set $x_{n}=(1-\alpha_n)y_{n-1}+\alpha_n z_{n-1}$,
  \item \textbf{Step 2}.
  Call the Oracle for computing $G(x_n,\xi_n)$ where $\EE[G(x_n,\xi_n)]=f'(x_n)$. Set 
  \begin{equation*}
   y_{n}=x_n-1/L_n G(x_n,\xi_n),
  \end{equation*}
  \begin{equation*}
z_n=z_{n-1}-\alpha_n^{-1}(x_n-y_n)  \end{equation*}
  \item \textbf{Step 3}.
  Set $n\rightarrow n+1$ and go to step 1.
 \end{itemize}

  We have 
  \begin{equation*}
      y_{n}=(I-1/L_n H_{n})x_n+\gamma_n\eps_{n},
  \end{equation*}
  and 
        \begin{eqnarray}
         z_n&=&z_{n-1}-\alpha_n^{-1}(x_n-y_n)  \nonumber \\
         &=&z_{n-1}-\alpha_n^{-1}[(1-\alpha_n)y_{n-1}+\alpha_n z_{n-1}-y_n]\nonumber\\
         &=&\alpha_n^{-1}y_n-\alpha_n^{-1}(1-\alpha_n)y_{n-1}. \nonumber
        \end{eqnarray}
      Thus
              \begin{eqnarray}
               x_{n}&=&(1-\alpha_n)y_{n-1}+\alpha_n z_{n-1}\nonumber \\
               &=&(1-\alpha_n)y_{n-1}+\alpha_n[\alpha_{n-1}^{-1}y_{n-1}-\alpha_{n-1}^{-1}(1-\alpha_{n-1})y_{n-2}]\nonumber\\
               &=&y_{n-1}+(1-\alpha_{n-1})\frac{\alpha_n}{\alpha_{n-1}}[y_{n-1}-y_{n-2}].\nonumber
              \end{eqnarray}
 These give the result for $\theta_n=y_n$.
\end{proof}

\subsection{Accelerated RDA method}
\begin{lemma}
 The algorithm AccRDA with step-sizes $\beta$ and $\alpha_n$ is equivalent to: 
 \begin{equation*}
  \theta_{n+1}=(I-\gamma_{n+1} H_{n+1})\theta_n+(1-\alpha_{n})\frac{\alpha_{n+1}}{\alpha_{n}}[I-\gamma_{n+1} H_{n+1}](\tn-\tnm)+\gamma_{n+1}\eps_{n+1},
 \end{equation*}
 with $\gamma_{n}=\frac{\alpha_{n}\theta_{n}}{L+\beta}$.
\end{lemma}
\begin{proof}
  We recall the general \textbf{Accelerated RDA method}:
 \begin{itemize}
  \item Let the initial points $w_0=v_0$, $A_0=0$, $\tilde g_0=0$ and the step-sizes $\{\alpha_n\}_{n\leq1}$ and $\{\beta_n\}_{n\leq1}$ be given. 
  
  Set $n=1$
  \item \textbf{Step 1}.
  Set $A_n=A_{n-1}+\alpha_n$ and $\theta_n=\frac{\alpha_n}{A_n}$.
  \item \textbf{Step 2}.
 Compute the query point $u_n=(1-\theta_n)w_{n-1}+\theta_nv_{n-1}$
 \item \textbf{Step 3}.
 Call the Oracle for computing $g_n=G(u_n,\xi_n)$ where $\EE[G(u_n,\xi_n)]=f'(u_n)$, and update the weighted average $\tilde g_n$
 \begin{equation*}
  \tilde g_n=(1-\theta_n)\tilde g_{n-1}+\theta_n g_n.
 \end{equation*}
 \item \textbf{Step 4}.
 Set $v_n=v_0-\frac{A_n}{L+\beta_n}\tilde g_n$.
 \item \textbf{Step 5}.
 Set $w_n=(1-\theta_n) w_{n-1}+\theta_n v_n$.

 \item \textbf{Step 6}.
  Set $n\rightarrow n+1$ and go to step 1.
 \end{itemize}

  First we have 
                \begin{eqnarray}
                 v_n&=&v_0-\frac{A_n}{L+\beta_n}\tilde g_n\nonumber \\
                 &=&v_0-\frac{A_n}{L+\beta_n}[(1-\theta_n)\tilde g_{n-1}+\theta_n g_n]\nonumber \\
                 &=&v_0-\frac{A_n}{L+\beta_n}[(1-\theta_n)\tilde g_{n-1}+\theta_n (H_{n+1}u_n+\eps_{n+1})]\nonumber \\
                 &=&v_0+(1-\theta_n)\frac{A_n(L+\beta_{n-1})}{(L+\beta_n)A_{n-1}}v_{n-1}-\frac{A_n}{L+\beta_n}\theta_n (H_{n+1}u_n+\eps_{n+1})]\nonumber \\
                 &=&v_0+(1-\theta_n)\frac{A_n(L+\beta_{n-1})}{(L+\beta_n)A_{n-1}}v_{n-1}-\frac{\alpha_n}{L+\beta_n} (H_{n+1}u_n+\eps_{n+1})].\nonumber                
                \end{eqnarray}
With $\beta_n=\beta$ we have $v_n=v_{n-1}-\frac{\alpha_n}{L+\beta} (H_{n+1}u_n+\eps_{n+1})]$ and
\begin{equation*}
 w_n=(I-\frac{\alpha_n\theta_n}{L+\beta} H_{n+1})u_n+\frac{\alpha_n\theta_n}{L+\beta} \eps_{n+1}.
\end{equation*}

Since $v_{n-1}=\theta_{n-1}^{-1}w_{n-1}-\theta_{n-1}^{-1}(1-\theta_{n-1})w_{n-2}$,
then 
\begin{equation*}
u_n=(1-\theta_n)w_{n-1}+\theta_n(\theta_{n-1}^{-1}w_{n-1}-\theta_{n-1}^{-1}(1-\theta_{n-1})w_{n-2}),
\end{equation*}    and 
\begin{equation*}
 u_n=w_{n-1}+\frac{\alpha_n A_{n-2}}{\alpha_{n-1}A_n}[w_{n-1}-w_{n-2}].
\end{equation*}           
\end{proof}

\section{Lower bound for stochastic optimization for least-squares}

\label{app:minimax}
In this section, we show a lower bound for optimization of quadratic functions with noisy access to gradients. We follow very closely the framework of~\citet{agarwal2010information} and use their notations.
The only difference with their Theorem 1 in the different choice of two functions $f_i^+$ and $f_i^-$, which we choose to be:
$$
f_i^\pm(x) = c_i ( x_i \pm \frac{r}{2} )^2,
$$
with a non-increasing sequence $(c_i)$ to be chosen later.
The function $g_\alpha$ that is optimized is thus:
$$
g_\alpha(x) = \frac{1}{d} \sum_{i=1}^d \big\{(\frac{1}{2} + \alpha_i \delta )f_i^+(x)  +  (\frac{1}{2} - \alpha_i \delta )f_i^-(x)  \big\}.
$$
This function is quadratic and its Hessian has eigenvalues equal to $2c_i/d$. Thus, its largest eigenvalue is $2c_1/d$, which we choose equal to $L$.

Noisy gradients are obtained by sampling $d$ independent Bernoulli random variables $b_i$, $i=1,\dots,d$,  with parameters $(\frac{1}{2} + \alpha_i \delta )$ and using
 the gradient of the random function $ \frac{1}{d} \sum_{i=1}^d \big\{ b_i f_i^+(x)  +  (1-b_i )f_i^-(x)  \big\}$. The variance of the random gradient is equal to
 $$
V =  \sum_{i=1}^d \frac{1}{d^2} {\rm var} \Big(
b_i \big[ c_i ( x_i + r/2) - c_i ( x_i - r /2) \big]
\Big) = \frac{1}{d^2} \sum_{i=1}^d c_i^2 r^2 ( 1/4 -   \delta^2 ).
 $$

The function $g_\alpha$ is minimized for $x = - \alpha \delta r$, and the discrepancy measure between two functions $g_\alpha$ and $g_\beta$ is greater than
$$\!\!\!\!
\frac{1}{d} \sum_{i=1}^d \bigg\{
 \inf_{x} \big\{ f_i^+(x)  +  f_i^-(x)  \big\} -   \inf_{x}   f_i^+(x)  -   \inf_{x}   f_i^-(x) \bigg\} 1_{\alpha_i \neq \beta_i}
\geqslant  \frac{1}{d} \sum_{i=1}^d \frac{3 c_i r^2 \delta^2}{4}  1_{\alpha_i \neq \beta_i}
\geqslant  \frac{1}{d} \frac{3 c_d r^2 \delta^2}{4}\Delta(\alpha, \beta).
$$
Since the vectors $\alpha,\beta \in \{-1,1\}^d$ are so that their Hamming distance $\Delta(\alpha, \beta) \geqslant d/4$ for $\alpha  \neq \beta$, we have a discrepancy measure greater than $ \frac{3 c_d r^2 \delta^2}{16} $.
Thus, for a an approximate optimality of $\varepsilon = \frac{  c_d r^2 \delta^2}{38}$, we have, following the proof of Theorem 1 (equation (29)) from~\citet{agarwal2010information}, for $N$ iterations of any method that accesses a random gradient, we have:
$$1/3 \geqslant  1-  2 \frac{16 N d \delta^2 + \log 2}{d \log (2/\sqrt{e})}.$$
Thus, for $d$ large, we get, up to constants, $\delta^2 \geqslant 1 / N$ and thus
$\varepsilon \geqslant \frac{r^2 c_d}{N}$.

For $c_1 = 2Ld$ and $c_i = L\sqrt{d}$ for the remaining ones, we get (up to constants):
$$
\varepsilon \geqslant \frac{V}{L } \frac{ \sqrt{d} }{N  }.
$$
This leads to the desired result for $N \leqslant d$.
\end{document}